\documentclass[letterpaper,journal]{IEEEtran}
\usepackage{amsmath,amsfonts}
\usepackage{algorithmic}
\usepackage{array}
\usepackage[caption=false,font=normalsize,labelfont=sf,textfont=sf]{subfig}
\usepackage{textcomp}
\usepackage{stfloats}
\usepackage{cuted}
\usepackage{xcolor}
\usepackage{url}
\usepackage{verbatim}
\usepackage{ragged2e}
\usepackage{bm}
\usepackage{tabularx}
\usepackage{multirow}
\usepackage{bbm}
\usepackage{amsmath}
\usepackage{graphicx}
\usepackage{booktabs}
\usepackage{amsbsy}
\usepackage{amssymb}
\usepackage{pifont}
\newcommand{\cmark}{\ding{51}}%
\newcommand{\xmark}{\ding{55}}%
\def\BibTeX{{\rm B\kern-.05em{\sc i\kern-.025em b}\kern-.08em
    T\kern-.1667em\lower.7ex\hbox{E}\kern-.125emX}}
\usepackage{balance}
\usepackage[colorlinks=true, linkcolor=blue, citecolor=blue, urlcolor=blue]{hyperref}
\usepackage{bookmark}
\usepackage[style=numeric-comp, backend=bibtex, sorting=none, minitems=2]{biblatex}
\begin{document}
\title{Learning Cross-View Semantic Priors for Single-Reference Unseen Object Pose Estimation}
\author{
Jiahong Chen, Jinghao Wang, Ziwen Wang, Zi Wang$^*$, Banglei Guan, \textit{Member, IEEE}, and Qifeng Yu
\thanks{Jiahong Chen, Jinghao Wang, Ziwen Wang, Zi Wang, Banglei Guan and Qifeng Yu are with the College of Aerospace Science and Engineering, National University of Defense Technology, Changsha 410073, China, and also with the Hunan Provincial Key Laboratory of Image Measurement and Vision Navigation, Changsha 410073, China (e-mail: chenjiahong@nudt.edu.cn; wangjinghao16@nudt.edu.cn; wangziwen24a@nudt.edu.cn; wangzi16@nudt.edu.cn; guanbanglei12@nudt.edu.cn; yuqifeng@nudt.edu.cn). Corresponding authors: Zi Wang.
}
}

\markboth{Submitted to IEEE Transactions on Image Processing}%
{How to Use the IEEEtran \LaTeX \ Templates}

\maketitle

\begin{abstract}
Single-reference unseen object 6D pose estimation reduces object onboarding by estimating poses of arbitrary novel objects from only one reference view.
Recent correspondence-based pipelines have achieved robust performance with vision foundation model (VFM) features.
However, they typically treat these features as intra-view descriptors, leaving dense visual-semantic cues, including appearance, structure, and context, insufficiently exchanged across views before geometric decoding.
Consequently, the decoded point features may lack joint semantic and geometric discriminability, making correspondence estimation still difficult in challenging cases.
Instead of processing features independently, we build the correspondence pipeline around an early cross-view semantic prior.
Specifically, cross-view semantic interaction (CVSI) enables dense query and reference VFM tokens to exchange semantic context and form a cross-view prior.
Nevertheless, direct CVSI may disturb the VFM token structure, while the resulting semantic prior still needs 3D representation consistency for rigid correspondence.
To make this CVSI prior reliable for 3D correspondence learning, we introduce two complementary training-time constraints: the intra-view structure preservation (IVSP) loss preserves the original intra-view token affinity structure during interaction, while the reference-anchored geometric consistency (RAGC) loss enforces spatial representation consistency of decoded point features.
The final pose is recovered from learned correspondences through weighted SVD.
We further construct a challenging view-pair protocol from the BOP Challenge datasets YCB-V and TUD-L to evaluate robustness in difficult matching scenarios.
Extensive experiments on six benchmarks under different view-pair settings show that our method achieves state-of-the-art performance while maintaining comparable inference speed.
Code, data, and models will be available at \href{https://chenjiahongbq.github.io/LCVSP}{\textcolor{magenta}{\textit{chenjiahongbq.github.io/LCVSP}}}.
\end{abstract}


\section{Introduction}
\IEEEPARstart{O}{bject} 6D pose estimation recovers the 3D rotation and translation of an object from visual observations, and is a fundamental capability for robotic manipulation~\cite{Kroemer_JMLR21_review_robot_learning,du2021vision,Efficient_Center,DTPose}, embodied perception~\cite{nie2020total3dunderstanding,huang2018cooperative}, and augmented reality~\cite{marchand2015pose,su2019arvr}.
Existing methods are commonly studied under instance-level, category-level, and unseen-object settings.
Instance-level~\cite{posecnn,su2022zebrapose,PAMI2022pvnet,Resolving_Symmetry,liu2025gdrnpp,Line-Based_6-DoF} and category-level methods often achieve strong performance~\cite{wang2019nocs,6D-ViT,li_NeurIPS21_equipose,chen2024secondpose,Ren_2026_CVPR}, but remain limited by supervision for specific objects or predefined categories, making new object onboarding costly~\cite{liu2024deep,thalhammer2024challenges}.
These limitations motivate unseen object pose estimation, which targets arbitrary novel objects beyond known instances and predefined categories~\cite{park2020latentfusion,labbe2023megapose,ornek2023foundpose,foundationposewen2024,lin2023sam6d,Shi_2026_CVPR}.

\begin{figure}[!t]
\refstepcounter{figure}
\centering
\includegraphics[width=\linewidth]{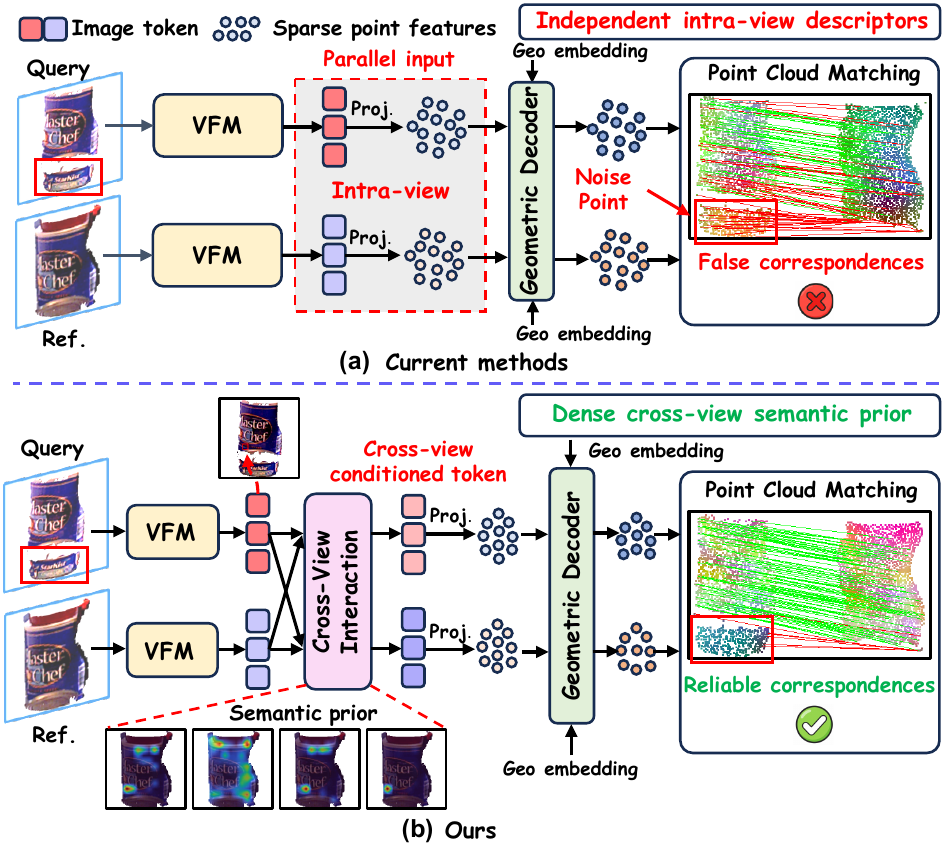}
\label{teaser}
\vspace{-16pt}
\begin{justify}
  \noindent\small \textbf{Fig. 1: Comparison with existing correspondence-based methods.} (a) Existing methods~\cite{lin2023sam6d,Liu_2025_CVPR,che2026cog,liu2026scalable} mainly use VFM features as intra-view descriptors for point cloud matching.
  (b) Our method performs cross-view semantic interaction on dense VFM tokens, forming a cross-view semantic prior for geometry-aware matching. 
\end{justify}
\end{figure}

To further reduce object onboarding, recent studies have moved toward the single-reference setting~\cite{nguyen2024nope,fan2023pope,corsetti2024oryon,Horyon,Liu_2025_CVPR,one2any2025,lee2025any6d,zuo2026coordar,che2026cog,kuang2025conceptpose,liu2026scalable}.
In this setting, only one RGB-D observation of the target object is available as the reference.
A common solution is to follow a correspondence-based formulation~\cite{lin2023sam6d,Liu_2025_CVPR,che2026cog,liu2026scalable}, where the relative pose is estimated from correspondences between the query and reference observations.
The central challenge is to make these correspondences reliable under large viewpoint changes and noisy observations.
Vision foundation models (VFMs)~\cite{oquab2023dinov2,simeoni2025dinov3} provide strong visual representations and have recently been adopted in single-reference pose estimation.
However, existing methods typically use VFM features as intra-view descriptors.
As illustrated in Fig.~\ref{teaser}(a), dense VFM feature maps are extracted independently from the query and reference images, projected onto sampled foreground point clouds, and then processed by a geometric decoder~\cite{qin2023geotransformer} for point matching and pose recovery.
This design has been effective for geometry-aware matching, but it does not fully condition dense visual-semantic cues across views before geometric decoding.
Appearance, part structure, and contextual relations encoded by VFM tokens are therefore weakened when they are used only as independent view-wise descriptors.
As a result, the decoded point features may lack joint semantic and geometric discriminability: they can encode local 3D consistency, but remain weak at distinguishing truly corresponding object regions from geometrically plausible yet semantically unrelated regions.

\begin{figure}[!t]
	\centering
    \refstepcounter{figure}
	\includegraphics[width=\linewidth]{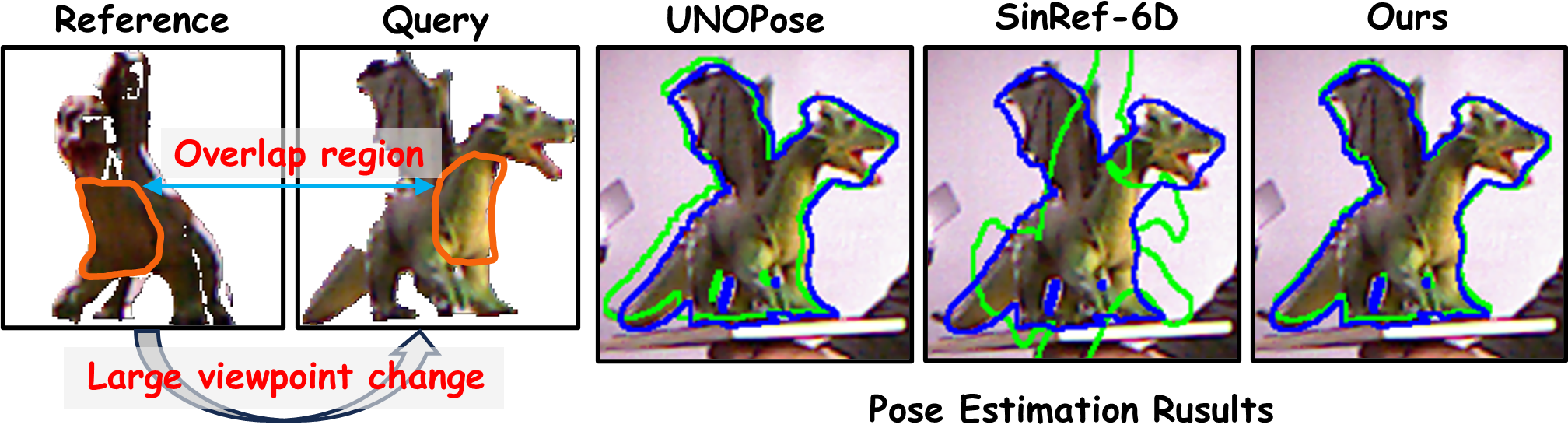}
    \label{fig:large_view}
    \vspace{-17pt}
    \begin{justify}
        \small \textbf{Fig. 2: A sample with large viewpoint changes.}
        Although the overlapping regions are very small, our method still achieves accurate pose estimation. 
        \textcolor{blue}{Blue} and \textcolor{green!60!black}{green} contours denote GT and estimated poses, respectively.
    \end{justify}
\end{figure}

This limitation becomes evident in two typical cases.
First, under noisy segmentation masks, as shown in Fig.~\ref{teaser}(a), the sampled foreground point cloud may contain nearby objects whose local geometry resembles the target.
If the decoder mainly relies on geometry at this stage, these distractor points can receive high matching salience and lead to false correspondences.
Second, under large viewpoint changes, the visible overlap between the query and reference observations becomes sparse and spatially ambiguous, as shown in Fig.~\ref{fig:large_view}.
In both cases, reliable matching requires a cross-view semantic prior before geometric decoding, so that point-level correspondence learning can be guided by reference-conditioned visual semantics rather than relying only on sparse geometric cues.

Motivated by this observation, we build the correspondence pipeline around an early cross-view semantic prior.
As shown in Fig.~\ref{teaser}(b), our method introduces cross-view semantic interaction (CVSI) at the dense VFM token level before projection and geometric decoding.
Instead of treating query and reference VFM features as independent intra-view descriptors, CVSI allows dense tokens from the two views to exchange semantic context and condition each other.
The resulting tokens form a visual semantic prior conditioned cross views.
When fused with geometry-aware embeddings in the decoder, this prior encourages the decoded point features to be both semantically selective and geometrically consistent for correspondence estimation.
However, learning such a prior is non-trivial because the query and reference observations are unposed, partially overlapping, and may contain mask-induced distractors.
Direct dense token interaction can therefore cause excessive feature mixing and over-smooth the spatial organization encoded by the VFM.
In addition, the learned prior should be translated into spatially consistent point features to benefit rigid 3D correspondence estimation.
To address these issues, we introduce two complementary training-time constraints that jointly make the CVSI prior reliable for 3D correspondence learning.
The intra-view structure preservation (IVSP) loss preserves the original intra-view token affinity structure during cross-view interaction, avoiding over-smoothed spatial organization.
The reference-anchored geometric consistency (RAGC) loss enforces spatial representation consistency of decoded point features in a shared reference frame, making the semantic prior compatible with rigid correspondence estimation.
The final pose is recovered from learned correspondences through weighted SVD.
Together, CVSI, IVSP, and RAGC are designed to improve correspondence reliability under large viewpoint changes and noisy masks.
To evaluate this capability, we further construct a challenging view pair protocol from the BOP Challenge datasets YCB-V~\cite{posecnn} and TUD-L~\cite{hodan2018bop}.
Extensive experiments on LM-O~\cite{brachmann2014learning}, TUD-L~\cite{hodan2018bop}, YCB-V~\cite{posecnn}, REAL275~\cite{wang2019nocs}, Toyota-Light~\cite{hodan2018bop}, and LINEMOD~\cite{hinterstoisser2012model} demonstrate that our method achieves state-of-the-art performance across different view pair settings while maintaining comparable inference speed.

The main contributions of this work are as follows:
\begin{itemize} 
    \item We identify the absence of an explicit cross-view semantic prior as a key limitation of existing correspondence-based methods, where VFM features are mainly used as intra-view descriptors before geometric decoding.

    \item We introduce cross-view semantic interaction (CVSI) at the dense VFM token level, forming an early cross-view semantic prior for subsequent geometry-aware point cloud matching.

    \item We introduce two complementary training-time constraints to make the CVSI prior reliable for 3D correspondence learning. IVSP preserves the intra-view token affinity structure during cross-view interaction, while RAGC enforces spatial representation consistency of decoded point features in a shared reference frame.

    \item We construct a challenging view-pair protocol from the BOP Challenge datasets YCB-V and TUD-L, focusing on robustness to noisy masks and large viewpoint changes in single-reference unseen object pose estimation.
\end{itemize}


\section{RELATED WORK}
\label{sec:related_work}

\subsection{Novel Object Pose Estimation}
\label{sec:rw_novel_pose}
Novel object pose estimation aims to recover the 6D pose of arbitrary objects beyond known instances and predefined categories.
One line of work addresses this setting by using object models, rendered templates, or generalizable matching at test time.
MegaPose~\cite{labbe2023megapose} follows a render-and-compare paradigm, where CAD-based renderings are retrieved and refined for pose estimation.
GigaPose~\cite{nguyen2024gigaPose} improves CAD-based novel object pose estimation with discriminative rendered templates and a compact correspondence formulation.
Recent methods further exploit vision foundation models~\cite{oquab2023dinov2,simeoni2025dinov3} and general segmentation priors~\cite{kirillov2023segment}.
SAM-6D~\cite{lin2023sam6d} combines SAM-based object proposal generation with semantic, appearance, and geometric matching, and formulates pose estimation as partial-to-partial point matching.

When CAD models are unavailable, another direction represents the target object using multiple reference observations~\cite{sun2022onepose,he2022onepose++,liu2022gen6d,he2022fs6d,foundationposewen2024}.
OnePose~\cite{sun2022onepose} reconstructs a sparse object model from a video scan using SfM, matches 2D query points to 3D SfM points, and estimates the object pose with PnP.
OnePose++~\cite{he2022onepose++} further removes the dependence on repeatable keypoint detection by using keypoint-free matching and reconstructing a semi-dense object point cloud.
Gen6D~\cite{liu2022gen6d} assumes several posed reference images and estimates pose through detection, viewpoint selection, and refinement.
FoundationPose~\cite{foundationposewen2024} unifies model-based and model-free pose estimation by using either a CAD model or a small set of reference images.
These methods improve generalization to novel objects, but they still typically rely on either CAD models or multiple reference views.

Recent works therefore move toward the more constrained single-reference setting~\cite{nguyen2024nope,corsetti2024oryon,zuo2026coordar,Horyon,Liu_2025_CVPR,lee2025any6d,one2any2025,che2026cog,kuang2025conceptpose,liu2026scalable}.
NOPE~\cite{nguyen2024nope} predicts pose-conditioned discriminative embeddings from a single reference image and estimates the query pose by matching them to generated viewpoint embeddings.
Oryon~\cite{corsetti2024oryon} and Horyon~\cite{Horyon} study open-vocabulary relative pose estimation, where a text prompt identifies the target object across two scenes and visual-language features are used for cross-scene matching and 3D registration.
One2Any~\cite{one2any2025} and CoordAR~\cite{zuo2026coordar} predict reference object coordinates from the query observation, converting single-reference pose estimation into alignment in a shared reference coordinate space.
UNOPose~\cite{Liu_2025_CVPR} casts single-reference pose estimation as point cloud registration by combining VFM semantic descriptors with an SE(3)-invariant geometric representation for matching.
COG~\cite{che2026cog} extends this correspondence-based formulation with confidence-aware optimal transport to obtain soft correspondences guided by point confidence.
Unlike UNOPose~\cite{Liu_2025_CVPR} and COG~\cite{che2026cog}, which mainly improve geometric decoding or correspondence assignment, we focus on feature construction before matching.
Our method performs cross-view semantic interaction at the dense VFM token level, forming a cross-view semantic prior for geometry-aware correspondence learning.

\subsection{Point Cloud Registration}
\label{sec:rw_registration}
Point cloud registration is closely related to single-reference unseen object pose estimation, as both require reliable correspondences between partial 3D observations.
Classical registration methods mainly rely on geometric matching.
ICP and its variants~\cite{ICP_tpami92,rusinkiewicz2001efficient} iteratively refine closest point associations, FPFH~\cite{rusu2009fast} describes local surface neighborhoods, and PPF~\cite{drost2010model} encodes oriented point pair relations for pose hypothesis generation.
These methods provide useful geometric baselines, but are sensitive to outliers, repeated local structures, and limited overlap.
Learning methods improve registration by estimating more discriminative correspondence cues.
Predator~\cite{huang2021predator} predicts overlap-aware features to focus matching on shared regions under low overlap.
GeoTransformer~\cite{qin2023geotransformer} encodes pairwise distances and angular relations to obtain geometric features that are invariant to rigid transformations for robust superpoint matching.

Unlike generic registration, single-reference unseen object pose estimation operates on mask-cropped partial RGB-D point clouds, where mask noise and large viewpoint changes introduce distractors and sparse overlap.
In such cases, geometry alone is often insufficient, and VFM cues~\cite{oquab2023dinov2,simeoni2025dinov3} can provide complementary appearance, layout, and contextual information.
This motivates a cross-view semantic prior before geometry-aware matching.

\subsection{Cross-View Correspondence Learning}
\label{sec:rw_cross_view}
Cross-view correspondence learning has been studied in image matching, object association, and visual geometry reasoning.
LoFTR~\cite{sun2021loftr} shows that transformer feature interaction can establish dense local correspondences without explicit keypoint detection.
O-MaMa~\cite{mur2025omama} formulates ego-exo object correspondence as mask matching, using DINOv2 features~\cite{oquab2023dinov2} and cross-attention for object level alignment.
V$^2$-SAM~\cite{pan2025v} adapts SAM2 to cross-view object correspondence by combining geometry-aware anchor prompts from DINOv3~\cite{simeoni2025dinov3} with visual prompts for appearance alignment.
VGGT~\cite{wang2025vggt} further demonstrates that cross-view spatial consistency can be encoded through unified visual geometry representations.

These studies show the value of cross-view interaction for transferring appearance, structure, and spatial cues, but mainly target image-level correspondence, mask association, or general visual geometry reasoning.
We study cross-view cues for single-reference unseen object pose estimation, where they must be grounded by geometric consistency for RGB-D correspondence learning and rigid pose recovery.

\section{Method}
\label{sec:method}

\subsection{Problem Formulation}
\label{sec:method_problem}

Given a query RGB-D image $[I^{q}\,|\,D^{q}]$ and a single reference RGB-D image $[I^{r}\,|\,D^{r}]$ of the same unseen rigid object, our goal is to estimate the relative object pose from the query view to the reference view, denoted by $\Delta T_{q\rightarrow r} \in SE(3)$. The reference image is accompanied by an object mask $M^{r}$ indicating the target object. For the query image, the object mask $M^{q}$ can be obtained either by a segmentation foundation model~\cite{kirillov2023segment} in the full pipeline or by ground-truth annotation when evaluating the pose estimator in isolation.

Using $M^{q}$ and $M^{r}$, we filter out background pixels, back-project the masked depth maps into 3D space, and sample the same number of foreground points to construct a query point cloud $\mathcal{P}^{q} = \{p_i^{q}\}_{i=1}^{N}$ and a reference point cloud $\mathcal{P}^{r} = \{p_i^{r}\}_{i=1}^{N}$, where $p_i^{q}, p_i^{r} \in \mathbb{R}^{3}$. Our method takes these sampled partial observations as input and predicts their relative rigid transformation $\Delta T_{q\rightarrow r}$. If the absolute pose of the reference object is available, the absolute pose of the query object can be recovered by $T^{q} = \Delta T_{q\rightarrow r}\, T^{r}$, where $T^{r}$ denotes the absolute pose of the reference object.

\subsection{Overview}
\label{sec:method_overview}
\begin{figure*}[t]
  \centering
  \refstepcounter{figure}
  \includegraphics[width=\linewidth]{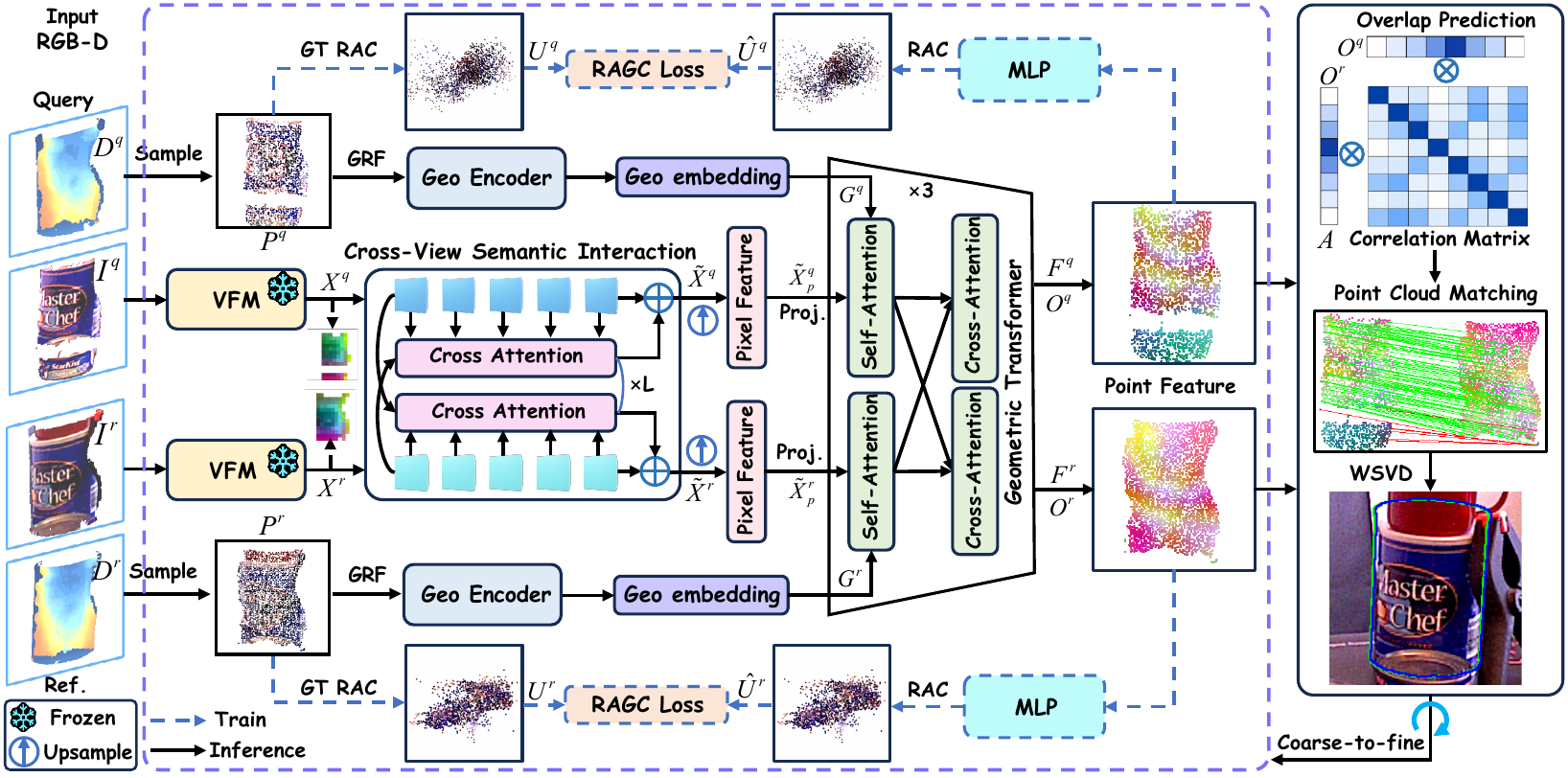}
  \label{fig:pipeline}
  \vspace{-13pt}
  \begin{justify}
    \small \textbf{Fig. 3: Overview of the proposed pipeline.} Given query and reference RGB-D observations after mask filtering, denoted by $[I^{q}\,|\,D^{q}]$ and $[I^{r}\,|\,D^{r}]$, we first extract image tokens $X^{q}$ and $X^{r}$ with a VFM~\cite{oquab2023dinov2,simeoni2025dinov3}. 
    The sampled point clouds are then transformed into a global reference frame (GRF)~\cite{Liu_2025_CVPR} and processed by a geometric encoder~\cite{qin2023geotransformer} to produce geometric features $G^{q}$ and $G^{r}$. Meanwhile, the image tokens exchange cross-view semantic information through $L$ CVSI blocks. 
    The resulting tokens, which encode a cross-view semantic prior, together with the geometric features are fed into the geometric transformer~\cite{qin2023geotransformer}, where semantic and geometric cues jointly guide decoding to obtain the final point features $F^{q}$ and $F^{r}$. 
  \end{justify}
\end{figure*}

As shown in Fig.~\ref{fig:pipeline}, our framework follows a coarse-to-fine correspondence pipeline. 
The coarse stage estimates an initial relative pose from sparse query and reference point clouds, while the fine stage repeats the pipeline with denser points after coarse alignment for pose refinement.
For clarity, we describe the main flow using the coarse stage as an example.

Given the query and reference RGB images, we first extract VFM tokens $\mathbf{X}^{q}$ and $\mathbf{X}^{r}$~\cite{oquab2023dinov2,simeoni2025dinov3}. 
Instead of using them as independent view-wise descriptors, we process them with $L$ CVSI blocks to obtain cross-view enhanced tokens $\tilde{\mathbf{X}}^{q}$ and $\tilde{\mathbf{X}}^{r}$.
These tokens are upsampled and projected onto the sampled point clouds as point-wise semantic features.
In parallel, the sampled point clouds are transformed into a pose-invariant global reference frame (GRF)~\cite{Liu_2025_CVPR} and encoded by a geometric encoder~\cite{qin2023geotransformer} to obtain geometric embeddings.

The point-wise semantic features and geometric embeddings are then fused by a geometric decoder~\cite{qin2023geotransformer}, producing correspondence features $\mathbf{F}^{q}$ and $\mathbf{F}^{r}$ together with overlap confidences $\mathbf{O}^{q}$ and $\mathbf{O}^{r}$. 
During training, the decoded features are additionally supervised by the RAGC loss through reference-anchored coordinate prediction, while the CVSI tokens are regularized by IVSP to preserve their intra-view structure.
At inference time, the features and overlap confidences form an overlap-aware correlation matrix $\mathbf{A}$, from which correspondences and pose hypotheses are generated following~\cite{Liu_2025_CVPR}. 
The best hypothesis initializes the fine stage, and the final relative pose is recovered from the refined correspondences by rigid registration:
\begin{equation}
\min_{\Delta R,\Delta t}
\sum_{(p_i^{q},p_j^{r})\in\mathcal{C}}
\left\|
\Delta R\,p_i^{q}+\Delta t-p_j^{r}
\right\|_{2}
\end{equation}
where $\mathcal{C}$ denotes the correspondence set extracted from $\mathbf{A}$.

The fine stage follows the same semantic interaction and geometric decoding pipeline, but differs in two aspects.
First, it samples denser point clouds $\mathcal{P}_{f}^{q}$ and $\mathcal{P}_{f}^{r}$ with $N_f>N_c$.
Second, after coarse alignment, its geometric encoding further concatenates local reference frame (LRF)~\cite{Liu_2025_CVPR} features and global position features encoded by a mini-PointNet~\cite{qi2017pointnet}, enabling more precise correspondence estimation and pose refinement.

\vspace{-10pt}
\subsection{Cross-view Semantic Interaction}
\label{sec:method_cvsi}
\begin{figure}[t]
  \centering
  \refstepcounter{figure}
  \includegraphics[width=\linewidth]{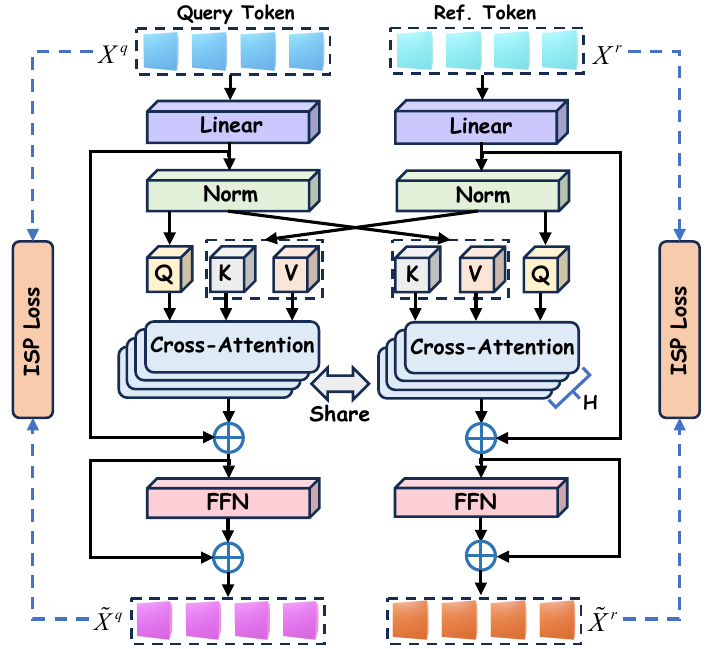}
  \label{fig:cvsi}
  \vspace{-14pt}
  \begin{justify}
    \small \textbf{Fig. 4: Illustration of cross-view semantic interaction.} The query and reference tokens are first projected to a lower-dimensional space, and then exchange semantic information through shared bidirectional cross-attention. 
  \end{justify}
\end{figure}

To further exploit cross-view semantic cues before point-level geometric decoding, we introduce cross-view semantic interaction (CVSI) to exchange semantic context between query and reference at the image token level.
As illustrated in Fig.~\ref{fig:cvsi}, this interaction conditions each view on the other, enabling dense tokens to highlight semantically related object regions and provide a cross-view prior for subsequent geometric point matching.

Given the semantic image tokens $\mathbf{X}^{q}$ and $\mathbf{X}^{r}$, we first project them to a lower-dimensional space through linear layers. Let $\bar{\mathbf{X}}^{q,l}$ and $\bar{\mathbf{X}}^{r,l}$ denote the resulting normalized low-dimensional features at the $l$-th block. 
We then compute the projected queries, keys, and values and apply shared bidirectional multi-head cross-attention. Taking the query branch as an example, the query vectors are computed from the query tokens, while the key and value vectors are computed from the reference tokens:
\begin{equation}
\mathbf{Q}^{q,l}=\bar{\mathbf{X}}^{q,l}\mathbf{W}_{Q}, \quad
\mathbf{K}^{r,l}=\bar{\mathbf{X}}^{r,l}\mathbf{W}_{K}, \quad
\mathbf{V}^{r,l}=\bar{\mathbf{X}}^{r,l}\mathbf{W}_{V}
\end{equation}
where $\mathbf{W}_{Q}$, $\mathbf{W}_{K}$, and $\mathbf{W}_{V}$ are learnable projection matrices.
The attended response at the $l$-th CVSI block is:
\begin{equation}
\Delta \mathbf{X}^{q,l}
=
\mathrm{softmax}
\left(
\frac{\mathbf{Q}^{q,l}{\mathbf{K}^{r,l}}^{\top}}{\sqrt{d}}
\right)\mathbf{V}^{r,l}
\end{equation}
The reference branch is updated in the same way through the shared cross-attention module, allowing semantic evidence to flow in both directions. 
The attended response is then injected back into the original token stream through residual updates and feed-forward refinement:
\begin{equation}
\mathbf{X}^{q,l+1}
=
\mathbf{X}^{q,l}
+ \Delta \mathbf{X}^{q,l}
+ \mathrm{FFN}\!\left(
\mathrm{LN}(\mathbf{X}^{q,l}+\Delta \mathbf{X}^{q,l})
\right)
\end{equation}
The reference update follows the same form. 
This residual design stabilizes cross-view interaction by suppressing semantic noise while preserving the informative content of the original tokens. 
After $L$ blocks, we obtain the cross-view enhanced tokens $\tilde{\mathbf{X}}^{q}=\mathbf{X}^{q,L}$ and $\tilde{\mathbf{X}}^{r}=\mathbf{X}^{r,L}$, which have already exchanged semantic information across views. 
These tokens are then projected to the sampled points as semantic features, allowing the geometric decoder to start from cross-view conditioned representations for point matching.

\subsection{Intra-view Structure Preservation}
\label{sec:method_isp}
As illustrated in Fig.~\ref{fig:cvsi}, to stabilize CVSI, we further introduce intra-view structure preservation (IVSP) to preserve the intrinsic intra-view organization of visual tokens during cross-view interaction. 

We therefore treat $\mathbf{X}^{q}$ and $\mathbf{X}^{r}$ as teacher tokens before CVSI, and $\tilde{\mathbf{X}}^{q}$ and $\tilde{\mathbf{X}}^{r}$ as student tokens after CVSI. 
For each view $v\in\{q,r\}$, let $M_v$ denote the number of tokens in that view. We first define the teacher similarity matrix from the pre-interaction tokens:
\begin{equation}
\mathbf{A}_{D}^{v}(i,j)
=
\cos\!\left(
\frac{\mathbf{X}_{i}^{v}}{\|\mathbf{X}_{i}^{v}\|_{2}},
\frac{\mathbf{X}_{j}^{v}}{\|\mathbf{X}_{j}^{v}\|_{2}}
\right)
\end{equation}
We then define the student similarity matrix from the post-interaction tokens. For the student branch, we keep only the nonnegative part of the cosine similarity, so that the supervision focuses on structurally relevant affinities. 
This also prevents negatively correlated regions in the post-interaction token space from being directly perturbed by the teacher signal:
\begin{equation}
\mathbf{A}_{S}^{v}(i,j)
=
\max\!\left(
\cos\!\left(
\frac{\tilde{\mathbf{X}}_{i}^{v}}{\|\tilde{\mathbf{X}}_{i}^{v}\|_{2}},
\frac{\tilde{\mathbf{X}}_{j}^{v}}{\|\tilde{\mathbf{X}}_{j}^{v}\|_{2}}
\right), 0
\right)
\end{equation}
Here, $i,j\in\{1,\dots,M_v\}$ index token pairs within the same view.
Although pretrained VFM features provide semantically meaningful correlations, their raw cosine similarities are not specifically optimized as direct supervision signals for the post-interaction token space~\cite{hamilton2022unsupervised}. 
We therefore center the teacher similarity matrix to remove the global cosine bias and retain relative intra-view structure:
\begin{equation}
\bar{\mathbf{A}}_{D}^{v}(i,j)
=
\mathbf{A}_{D}^{v}(i,j)
- \frac{1}{M_v^2}\sum_{m,n}\mathbf{A}_{D}^{v}(m,n)
\end{equation}
where $m,n\in\{1,\dots,M_v\}$ enumerate all token pairs in view $v$.
We then regularize the intra-view structure for each view by:
\begin{equation}
\mathcal{L}_{\text{IVSP}}^{v}
=
\frac{1}{M_v^2}
\sum_{i,j}
-
\bar{\mathbf{A}}_{D}^{v}(i,j)\mathbf{A}_{S}^{v}(i,j)
\end{equation}
The final token structure preservation loss averages the two views:
\begin{equation}
\mathcal{L}_{\text{IVSP}}
=
\frac{1}{2}
\left(
\mathcal{L}_{\text{IVSP}}^{q}
+
\mathcal{L}_{\text{IVSP}}^{r}
\right)
\end{equation} 
This regularization preserves the structural topology of the original token manifold, so that CVSI learns to reweight semantically relevant regions across views rather than collapsing unrelated tokens into an over-smoothed representation.

\subsection{Reference-anchored Geometric Consistency}
\label{sec:method_ragc}

To make the CVSI prior beneficial for 3D matching, we impose a reference-anchored geometric consistency (RAGC) loss on the decoded point features, inspired by explicit 3D representation learning~\cite{wang2025vggt}.
RAGC supervises auxiliary coordinate prediction in a shared reference frame, encouraging the features to couple cross-view semantic compatibility with geometric consistency for reliable correspondence estimation.

To stabilize network training, we adopt the reference-anchored coordinate (RAC) normalization strategy of~\cite{one2any2025} and define an RAC space from the reference point cloud $\mathcal{P}^{r}=\{p_i^r\}_{i=1}^{N}$.  
Let $\mathbf{c}^{r}$ denote the center of $\mathcal{P}^{r}$, and let $w^{r} = \max_{i,j}\|p_i^{r} - p_j^{r}\|_{2}$ denote the maximum distance between reference points.
Following this normalization strategy, the scale is given by $s^{r} = 1 / (1.1\,w^{r})$ and the translation by $\mathbf{t}^{r} = -s^{r}\mathbf{c}^{r}$, yielding the reference-anchored transform $T_{\mathrm{rac}}^{r}(p) = s^{r}p + \mathbf{t}^{r}$, which maps the reference object into a normalized RAC space.
Importantly, this transform is computed from the original unsampled reference point cloud rather than the sampled coarse or fine points, so that the target coordinate system remains stable across different sampling stages.
Based on $T_{\mathrm{rac}}^{r}$, we define the RAC targets for the reference and query points as $\mathbf{U}^{r}=\{u_j^{r}\}_{j=1}^{N}$ and $\mathbf{U}^{q}=\{u_i^{q}\}_{i=1}^{N}$, respectively:
\begin{equation}
u_j^{r} = T_{\mathrm{rac}}^{r}(p_j^{r}), \qquad
u_i^{q} = T_{\mathrm{rac}}^{r}(R^{*}p_i^{q}+t^{*})
\end{equation}
where $(R^{*},t^{*})$ is the ground-truth relative pose. 
In this way, reference points and ground-truth aligned query points are represented in the same reference-anchored coordinate system. 

To inject this constraint into feature learning, we feed the fused point features $\mathbf{F}^{q}$ and $\mathbf{F}^{r}$ into a multi-layer MLP. Concretely, the MLP predicts RAC coordinates by $\hat{\mathbf{U}}^{q}=f(\mathbf{F}^{q})$ and $\hat{\mathbf{U}}^{r}=f(\mathbf{F}^{r})$, where $f(\cdot)$ denotes the mapping function. Let $\rho(\cdot,\cdot)$ denote the SmoothL1 loss. We then supervise the predicted coordinates $\hat{\mathbf{U}}^{q}=\{\hat{u}_i^{q}\}_{i=1}^{N}$ and $\hat{\mathbf{U}}^{r}=\{\hat{u}_j^{r}\}_{j=1}^{N}$ with:
\begin{equation}
\mathcal{L}_{\text{RAGC}}
=
\frac{1}{2}
\left(
\frac{1}{N}\sum_{i=1}^{N}\rho(\hat{u}_{i}^{q},u_{i}^{q})
+
\frac{1}{N}\sum_{j=1}^{N}\rho(\hat{u}_{j}^{r},u_{j}^{r})
\right)
\end{equation}
This loss turns RAC into a geometric consensus signal for the fused representation. 
By anchoring the decoded point features to a common reference geometry, it complements the semantic compatibility learned by CVSI and improves the stability and discriminability of cross-view correspondence estimation.

\subsection{Training objective}
\label{sec:method_training}

Our training objective combines overlap-aware correspondence supervision with two training-time regularization terms: RAGC for reference-anchored geometric consistency and IVSP for intra-view token structure preservation.

For correspondence learning, we adopt a coarse-to-fine supervision strategy following recent RGB-D matching pipelines~\cite{Liu_2025_CVPR}. 
The matching term $\mathcal{L}_{A}$ uses the InfoNCE~\cite{oord2018representation} loss to supervise the overlap-aware correlation matrix, while the overlap term $\mathcal{L}_{o}$ supervises the predicted overlap confidences with the weighted binary cross-entropy (WBCE) loss. 
For the $t$-th geometric transformer block, both terms are applied at the coarse and fine stages:
\begin{equation}
\mathcal{L}_{\text{match}}
=
\sum_{t=1}^{3}\bigl(\mathcal{L}_{A}^{c,t}+\mathcal{L}_{o}^{c,t}\bigr)
+
\sum_{t=1}^{3}\bigl(\mathcal{L}_{A}^{f,t}+\mathcal{L}_{o}^{f,t}\bigr)
\end{equation}
where the superscripts $c$ and $f$ denote the coarse and fine stages, respectively.
The proposed auxiliary losses are also applied at both stages. 
The full training objective is therefore
\begin{equation}
\mathcal{L}
=
\mathcal{L}_{\text{match}}
+
\sum_{s\in\{c,f\}}
\left(
\lambda_{\text{RAGC}}\,\mathcal{L}_{\text{RAGC}}^{s}
+
\lambda_{\text{IVSP}}\,\mathcal{L}_{\text{IVSP}}^{s}
\right)
\end{equation}
where $\lambda_{\text{RAGC}}$ and $\lambda_{\text{IVSP}}$ balance the contribution of the geometric and structural regularization terms.

\section{EXPERIMENTS}
\label{sec:experiments}

\subsection{Experimental Setup}
\label{sec:exp_setup}
\textbf{Implementation Details.} 
Our method is implemented in PyTorch~\cite{paszke2019pytorch}. For image encoding, we use a DINOv3~\cite{simeoni2025dinov3} pretrained ViT-Base backbone~\cite{dosovitskiy2020image}. 
Following~\cite{Liu_2025_CVPR,che2026cog}, the network is trained on the standard MegaPose synthetic dataset~\cite{labbe2023megapose} for the BOP unseen object pose estimation track~\cite{hodan2024bop}. 
We train the network for 440K steps with a batch size of 8 on the RTX 4090 GPU. Optimization is performed with Adam~\cite{KingmaB_adam_iclr15} and cosine annealing~\cite{loshchilov-ICLR17SGDR}, using a base learning rate of $10^{-4}$. 
Unless otherwise specified, the number of CVSI blocks is set to $L=3$, the number of attention heads is set to $H=8$, and the loss weights are fixed to $\lambda_{\text{RAGC}}=1$ and $\lambda_{\text{IVSP}}=1$. In the coarse stage, we sample $N_c=196$ points and generate 300 pose hypotheses. In the fine stage, we increase the number of sampled points to $N_f=2048$.

\textbf{Datasets.} 
We conduct experiments on six benchmark datasets: LM-O~\cite{brachmann2014learning}, TUD-L~\cite{hodan2018bop}, YCB-V~\cite{posecnn}, Real275~\cite{wang2019nocs}, Toyota-Light~\cite{hodan2018bop}, and LINEMOD~\cite{hinterstoisser2012model}. 
LM-O, TUD-L, and YCB-V follow the predefined view-pair protocol of~\cite{Liu_2025_CVPR}, where SAM masks~\cite{kirillov2023segment} introduce realistic segmentation noise and the scenes cover clutter, occlusion, lighting variation, and sensor noise. 
Real275 and Toyota-Light are evaluated under the Oryon protocol~\cite{corsetti2024oryon}, with 2,000 reference-query pairs and ground-truth masks for each test set, covering diverse indoor objects and challenging illumination conditions. 
To further assess sensitivity to the reference view, we also evaluate the first-frame single-reference setting on YCB-V~\cite{posecnn} and LINEMOD~\cite{hinterstoisser2012model}, following~\cite{foundationposewen2024,one2any2025}, where LINEMOD provides large viewpoint changes across object sequences.

\textbf{Challenging Benchmark.}
We further construct a challenging view-pair benchmark based on YCB-V~\cite{posecnn} and TUD-L~\cite{hodan2018bop}.
For each query object in the test split, we randomly select a reference view from the training or validation split and constrain the relative rotation angle to the range of 60 to 90 degrees.
For the object masks, we start from the segmentation results of UNOPose~\cite{Liu_2025_CVPR} and further inject noise by applying random morphological dilation.

\textbf{Evaluation Metrics.} 
For LM-O, TUD-L, and YCB-V under the predefined view-pair protocol, we follow the BOP evaluation protocol and report AR$_{\text{BOP}}$, averaged over VSD, MSSD, and MSPD~\cite{hodan2018bop}. 
For Real275, Toyota-Light, and LINEMOD, we report ADD(-S) at $0.1d$ with ADD-S for symmetric objects~\cite{hinterstoisser2012model,hodavn2016evaluation}.
For YCB-V under the first-frame reference setting, we report AUC of ADD and ADD-S following PoseCNN~\cite{posecnn}. 

\subsection{Comparison with State-of-the-Art Methods}
\label{sec:exp_sota}
\begin{table*}[!t]
	\centering
    \refstepcounter{table}
    \begin{justify}
    \small Table \Roman{table}: Pose estimation results on LM-O, TUD-L, and YCB-V under the view-pair setting of UNOPose~\cite{Liu_2025_CVPR}. Object masks are obtained by SAM~\cite{kirillov2023segment}. The mean Average Recall (AR) of the BOP metric and the average time (s) per image are reported. The runtime includes all instances from the SAM proposals.
    \end{justify}
    \label{tab:sota_1}
    \setlength{\tabcolsep}{6.0pt}
\begin{tabular}{@{}lccccccc@{}}
        \toprule
        \multirow{2}{*}{Method} & \multirow{2}{*}{Modality} & \multirow{2}{*}{Reference}
        & \multicolumn{4}{c}{AR$_{\text{BOP}}$ (\%)} & \multirow{2}{*}{Time (s) $\downarrow$} \\
        \cmidrule(lr){4-7}
        & & & LM-O~\cite{brachmann2014learning} $\uparrow$ & TUD-L~\cite{hodan2018bop} $\uparrow$ & YCB-V~\cite{posecnn} $\uparrow$ & Mean $\uparrow$ & \\
        \midrule
        FPFH+RANSAC~\cite{rusu2009fast} & D & Image & 31.0 & 31.0 & 50.0 & 37.3 & 6.38 \\
        FPFH+MAC~\cite{zhang20233d} & D & Image     & 22.5 & 22.1 & 49.6 & 31.4 & 136.94 \\
        PPF~\cite{drost2010model} & D & Image        & 29.7 & 14.8 & 38.3 & 27.6 & 11.79 \\
        PPF\_3D\_ICP~\cite{drost2010model} & D & Image  & 44.7 & 29.1 & 66.8 & 46.9 & 14.27 \\
        FCGF+RANSAC~\cite{choy2019fcgf} & D & Image & 38.9 & 59.0 & 57.6 & 51.8 & 10.96 \\
        FCGF+MAC~\cite{zhang20233d} & D & Image     & 33.9 & 48.3 & 51.0 & 44.4 & 60.53 \\
        UTOPIC \cite{chen2022utopic} & D & Image    & 13.7 & 35.4 & 10.5  & 19.9 & \textbf{4.00} \\
        GeDi~\cite{poiesi2022gedi}   & D & Image      & 42.8 & 67.3 & 60.6 & 56.9 & 48.89 \\
        \midrule
        FreeZe~\cite{caraffa2024freeze}& RGB-D & Image& 45.5 & 68.3 & 65.5 & 59.8 & 52.96 \\
        SAM-6D~\cite{lin2023sam6d}  & RGB-D & Posed Image & 54.5 & 29.7  &  68.1 & 50.8 & 4.21  \\
        SinRef-6D~\cite{liu2026scalable}  & RGB-D & Posed Image & 51.1 & 34.0  &  73.9 & 53.0 & 4.231  \\
        CoordAR~\cite{zuo2026coordar}  & RGB-D & Image & 46.7 & 52.1  &  67.3 & 55.4 & 6.063  \\
        UNOPose~\cite{Liu_2025_CVPR} & RGB-D & Image & 58.7 & 71.0 & \underline{83.1} & 70.9 & \underline{4.179} \\
		COG~\cite{che2026cog} & RGB-D & Image & \underline{60.8} & \underline{80.0} & 80.5 & \underline{73.8} & 4.324 \\
        \midrule
        \textbf{Ours} &  RGB-D & Image  & \textbf{61.2} & \textbf{82.0} & \textbf{86.7} & \textbf{76.6} &  4.232 \\
        \bottomrule
\end{tabular}
\end{table*}

Since the checkpoint of the best-performing (supervised version) COG model~\cite{che2026cog} is not publicly available, we report its results only in Table~\ref{tab:sota_1} using the numbers from the original paper.

\textbf{Comparison under the Predefined View Pair Protocol.} 
Table~\ref{tab:sota_1} compares different methods under the predefined view-pair protocol of UNOPose~\cite{Liu_2025_CVPR}.
Pure point cloud registration methods are limited in this setting, with the best one reaching only 56.9\% mean AR$_{\text{BOP}}$.
This indicates that geometry alone is insufficient for mask-cropped partial point clouds with distractors and limited overlap.
RGB-D single-reference methods perform better by combining visual and geometric information, with the strongest baselines UNOPose~\cite{Liu_2025_CVPR} and COG~\cite{che2026cog} reaching 70.9\% and 73.8\% mean AR$_{\text{BOP}}$, respectively.
Our method achieves the best performance on all three datasets, reaching 76.6\% mean AR$_{\text{BOP}}$ and improving over UNOPose and COG by 5.7\% and 2.8\%, respectively.
Compared with the strongest baseline on each dataset, our gains are 0.4\% on LM-O, 2.0\% on TUD-L, and 3.6\% on YCB-V.

\textbf{Runtime Analysis}. Due to differences in hardware from the original papers, we re-evaluate the inference time of SinRef-6D~\cite{liu2026scalable}, UNOPose~\cite{Liu_2025_CVPR}, COG~\cite{che2026cog}, and CoordAR~\cite{zuo2026coordar} on a single RTX 4090 GPU. Our method runs at 4.232\,s per image, including 2.68s for
segmentation (taken from UNOPose) and 1.552s for pose estimation. 
Compared with UNOPose, it introduces only 0.053\,s additional latency, while improving the average AR by 5.7\%.
\begin{table}[ht]
	\centering
    \refstepcounter{table}
    \begin{justify}
    \small Table \Roman{table}: pose estimation results on Real275~\cite{wang2019nocs} and Toyota-Light~\cite{hodan2018bop} under the view-pair protocol of Oryon~\cite{corsetti2024oryon} and Horyon~\cite{Horyon}, where ground truth object masks are used. 
    We compare RGB and RGB-D methods with only a single reference view.
    \end{justify}
    \label{tab:sota_2}
	\setlength{\tabcolsep}{3.6pt}
	\begin{tabular}{lccccc}
    \toprule
       \multirow{2}{*}{Methods}  & \multirow{2}{*}{Modality} & \multicolumn{2}{c}{Real275~\cite{wang2019nocs}} & \multicolumn{2}{c}{Toyota-Light~\cite{hodan2018bop}} \\
      \cmidrule(lr){3-6}
      &  & AR $\uparrow$  & ADD(-S) $\uparrow$ & AR $\uparrow$    & ADD(-S) $\uparrow$  \\
        \midrule
        PoseDiffusion \cite{wang2023posediffusion} & RGB & 9.2  & 0.8 & 7.8   & 1.2\\
        RelPose++ \cite{lin2023relpose++} & RGB  & 22.8 & 11.9 & 30.9  & 11.6 \\
        LatentFusion \cite{park2020latentfusion} & RGB &22.6  & 9.6 & 28.2  & 10.2 \\
        \midrule
        SIFT \cite{lowe1999object}& RGB-D& 34.1  & 16.4 & 30.3  & 14.1 \\
        ObjectMatch \cite{gumeli2023objectmatch}& RGB-D& 26.0 & 13.4 &9.8  & 5.4\\
        Oryon \cite{corsetti2024oryon}  & RGB-D& 46.5 & 34.9 & 34.1 & 22.9\\
		One2Any~\cite{one2any2025} & RGB-D & 54.9 &  41.0 & 42.0 & 34.6 \\
        Horyon~\cite{Horyon} & RGB-D & 57.9 &  51.6 & 33.0 & 25.1 \\
        Any6D~\cite{lee2025any6d} & RGB-D & 51.0 &  53.5 & 43.3 & 32.2 \\
        UNOPose~\cite{Liu_2025_CVPR} & RGB-D & \underline{77.9} &  \underline{84.4} & \underline{74.9} & 73.2 \\
        SinRef-6D~\cite{liu2026scalable} & RGB-D & 74.4 &  81.1 & 66.7 & 67.1 \\
        ConceptPose~\cite{kuang2025conceptpose} & RGB-D & 60.4 &  71.5 & 51.6 & 55.0 \\
        CoordAR~\cite{zuo2026coordar} & RGB-D & 71.0 &  82.2 & 62.5 & \textbf{82.6} \\
        \midrule
        Ours & RGB-D & \textbf{78.1} &  \textbf{86.2} & \textbf{78.4} & \underline{80.0} \\
        \bottomrule
    \end{tabular}
\end{table}

Table~\ref{tab:sota_2} reports the results on Real275 and Toyota-Light under the view-pair protocol of Oryon~\cite{corsetti2024oryon} and Horyon~\cite{Horyon}.
RGB-only methods remain clearly behind RGB-D methods, showing that depth is important for accurate relative pose estimation.
Among RGB-D single-reference methods, the strongest baseline UNOPose~\cite{Liu_2025_CVPR} reaches 77.9\% AR and 84.4\% ADD(-S) on Real275, and 74.9\% AR and 73.2\% ADD(-S) on Toyota-Light.
Our method achieves the best AR on both datasets and the best ADD(-S) on Real275.
Compared with UNOPose, it improves AR and ADD(-S) from 77.9\% to 78.1\% and from 84.4\% to 86.2\% on Real275, and from 74.9\% to 78.4\% and from 73.2\% to 80.0\% on Toyota-Light.
On Toyota-Light, our ADD(-S) is second only to CoordAR~\cite{zuo2026coordar}, while our AR remains substantially higher.

\textbf{Comparison under the One-Reference Per Object Setting.} To evaluate the robustness of our method to the reference view, we further fix one reference image for each object and test under a stricter single-reference protocol on YCB-V~\cite{posecnn} and LINEMOD~\cite{hinterstoisser2012model}.
Table~\ref{tab:sota_4} reports the results on YCB-V~\cite{posecnn} under the first-frame reference setting.
FoundationPose~\cite{foundationposewen2024} achieves the best overall result with 16 CAD-rendered references, but drops from 91.5\% to 76.1\% AUC of ADD when using a single CAD reference.
Under the single-image reference setting, our method achieves the best AUC of ADD at 88.1\%, outperforming One2Any~\cite{one2any2025}, UNOPose~\cite{Liu_2025_CVPR}, and CoordAR~\cite{zuo2026coordar} by 3.7\%, 5.3\%, and 9.6\%, respectively.
For AUC of ADD-S, our method obtains 95.7\%, which is close to UNOPose at 96.0\% and higher than the other single-reference baselines.
\begin{table}[ht]
	\centering
    \refstepcounter{table}
    \begin{justify}
    \small Table \Roman{table}: Pose estimation results on on YCB-V \cite{posecnn} dataset. 
    We compare with point cloud registration methods, multi-view methods, and single-view methods. 
    Predator \cite{huang2021predator}, LoFTR \cite{sun2021loftr} and FS6D \cite{he2022fs6d} are fine-tuned on the YCB-Video dataset. 
    Results of multi-view methods are adopted from \cite{foundationposewen2024}. 
    For single-view methods, we provide the first image in the test set as the reference.
    \end{justify}
    \label{tab:sota_4}
	\setlength{\tabcolsep}{4.3pt}
	\begin{tabular}{@{}lccc@{}}
    \toprule
        Method & Ref. Images & AUC of ADD-S $\uparrow$ & AUC of ADD $\uparrow$ \\
        \midrule
        Predator~\cite{huang2021predator} & 16 & 71.0 & 24.3 \\
        LoFTR~\cite{sun2021loftr} & 16 & 52.5 & 26.2 \\
        FS6D~\cite{he2022fs6d} & 16 & 88.4 & 42.1 \\
        FoundationPose~\cite{foundationposewen2024} & 16 - CAD & \textbf{97.4} & \textbf{91.5} \\
        \midrule
        FoundationPose~\cite{foundationposewen2024} & 1 - CAD & 90.4 & 76.1 \\
        Oryon~\cite{corsetti2024oryon} & 1 & 13.3 & 7.4 \\
        NOPE~\cite{nguyen2024nope} & 1 + GT trans & 86.0 & 25.1 \\
        One2Any~\cite{one2any2025} & 1 & 93.7 & \underline{84.4} \\
        SinRef-6D~\cite{liu2026scalable} & 1 & 93.9 & 78.4 \\
        CoordAR~\cite{zuo2026coordar} & 1 & 95.5 & 78.5 \\
        UNOPose~\cite{Liu_2025_CVPR} & 1 & \textbf{96.0} & 82.8 \\
		\midrule
		Ours & 1 & \underline{95.7} & \textbf{88.1} \\
        \bottomrule
    \end{tabular}
\end{table}

\begin{table*}[!t]
	\centering
    \refstepcounter{table}
    \begin{justify}
    \small Table \Roman{table}: Pose estimation results on LINEMOD~\cite{hinterstoisser2012model} dataset. 
    The upper block reports multi-reference methods, and the lower block reports single-reference methods. 
    For all single-reference image methods, the first view is used as the reference. 
    We report the recall of ADD(-S)-0.1d metric. 
    Results of multi-view methods are taken from FoundationPose~\cite{foundationposewen2024}. 
    \end{justify}
    \label{tab:sota_3}
    \setlength{\tabcolsep}{2.5pt}
\begin{tabular}{lc c c c c c c c c c c c c c c c}
\toprule
Methods & Modality  & Ref. Images & \textbf{ape} & \textbf{benchvise} & \textbf{cam} & \textbf{can} & \textbf{cat} &  \textbf{driller} & \textbf{duck} & \textbf{eggbox} & \textbf{glue} & \textbf{holepuncher} & \textbf{iron} & \textbf{lamp} & \textbf{phone} & \textbf{Mean} \\
\midrule
OnePose \cite{sun2022onepose} & RGB & 200 & 11.8 & 92.6 & 88.1 & 77.2 & 47.9 & 74.5 & 34.2 & 71.3 & 37.5 & 54.9 & 89.2 & 87.6 & 60.6 & 63.6 \\
OnePose++ \cite{he2022onepose++} & RGB  & 200 & 31.2 & \textbf{97.3} & 88.0 & \textbf{89.8} & 70.4 & \textbf{92.5} & 42.3 & \textbf{99.7} & 48.0 & 69.7 & \textbf{97.4} & \textbf{97.8} & 76.0 & 76.9 \\
LatentFusion \cite{park2020latentfusion} & RGB-D & 16 &  \textbf{88.0} & 92.4 & 74.4 & 88.8 & 94.5 & 91.7 & 68.1 & 96.3 & 49.4 & 82.1 & 74.6 & 94.7 & 91.5 & 83.6 \\
FS6D \cite{he2022fs6d} + ICP & RGB-D & 16 & 78.0 & 88.5 & \textbf{91.0} & 89.5 & \textbf{97.5} & 92.0 & \textbf{75.5} & 99.5 & \textbf{99.5} & \textbf{96.0} & 87.5 & 97.0 &\textbf{97.5} & \textbf{91.5} \\
\midrule 
FoundationPose \cite{foundationposewen2024}& RGB-D & 1-CAD & 36.5 & 55.5 & \textbf{84.2} & 71.7 &  65.3 & 16.3  & 49.8 & 42.6 & 64.8 & 52.7 & 20.7 & 15.8 & 51.7 & 48.3\\
NOPE \cite{nguyen2024nope} & RGB & 1 + GT trans & 2.0 & 4.5 &  2.5 & 2.2 & 0.7 & 4.7& 0.5 &  \textbf{100.0} & 79.4 & 2.9  & 4.5 & 4.2 & 3.9 &16.3\\
Oryon \cite{corsetti2024oryon} &  RGB-D & 1 & 1.2 & 1.3 & 3.9 & 0.8 & 12.7  & 8.5 & 0.8 & 63.2 & 18.4 & 1.6 & 0.6 &2.9 & 11.7 & 9.8 \\
One2Any~\cite{one2any2025} & RGB-D  & 1 & 33.1& 15.7 & 72.7 & 37.0 & 66.2  & 68.2 & 35.8 & \textbf{100.0} & \textbf{99.9} &  42.0 & 28.2 & 31.9 & 53.2 & 52.6 \\
UNOPose~\cite{Liu_2025_CVPR} & RGB-D  & 1 & 44.6& 54.9 & 80.2 & 47.1 & 80.7  & 89.4 & 45.2 & 99.2 & 97.2 &  75.3 & 51.8 & 64.0 & 76.6 & 69.7 \\
CoordAR \cite{zuo2026coordar} & RGB-D  & 1 & 45.6 & 76.9 & 70.7 & 77.3 & 88.1 & \textbf{96.5} & 50.2 & 97.0 & 99.8 & 67.5 & 52.7 & 91.4 & 61.2 & 75.0 \\
SinRef-6D \cite{liu2026scalable} & RGB-D  & 1 & \textbf{49.4} & 82.1 & 63.2 & 58.1 & \textbf{88.3} & 76.9 & 53.7 & 99.7 & 83.5 & 46.4 & \textbf{86.5} & \textbf{99.6} & 85.8 & 74.9 \\
\midrule
Ours & RGB-D  & 1 & 43.0 & \textbf{92.6} & 82.8 & \textbf{81.4} & 87.5  & 78.9 & \textbf{63.2} & 99.7 & 75.4 &  \textbf{77.4} & 66.1 & 72.6 & \textbf{85.9} & \textbf{77.4} \\
\bottomrule
\end{tabular}
\end{table*}

\begin{figure*}[!t]
  \centering
  \refstepcounter{figure}  \includegraphics[width=0.96\linewidth]{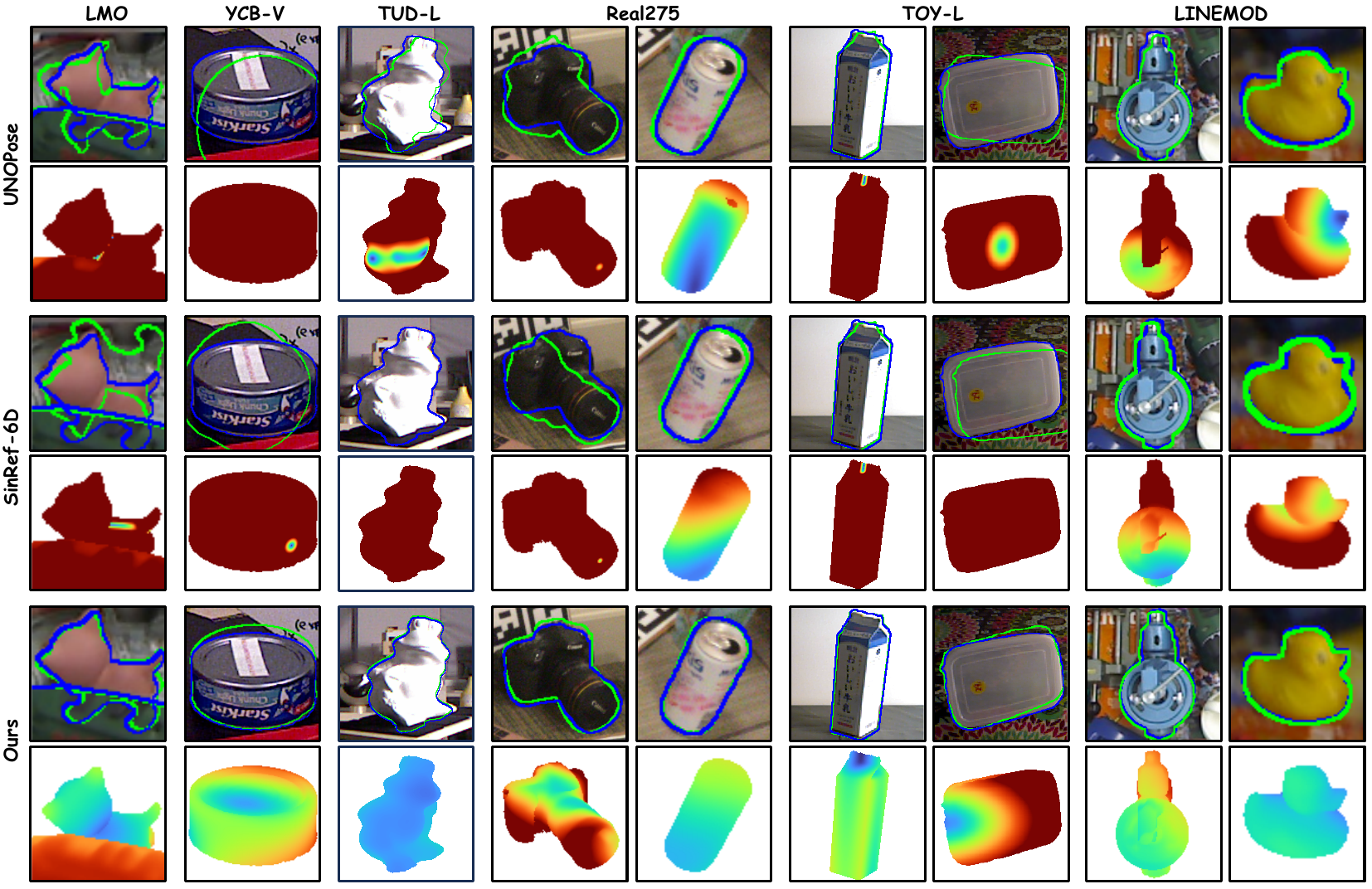}
  \label{fig:comparison_vis}
  \vspace{-6pt}
  \begin{justify}
    \small \textbf{Fig. 5: Qualitative comparison on six datasets.} We visualize the pose estimation results of UNOPose~\cite{Liu_2025_CVPR}, SinRef-6D~\cite{liu2026scalable}, and our method. \textcolor{blue}{Blue} and \textcolor{green!60!black}{green} contours denote GT and estimated poses, respectively. For clearer visualization, we also show the depth error heatmap of each detected object with respect to the ground-truth pose, namely the distance between each 3D point in the ground-truth depth map and its transformed position under the predicted pose (legend: 0~cm \includegraphics[height=1.8mm]{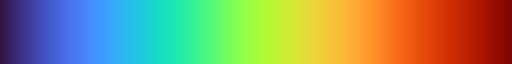} 10~mm). 
    More results can be found in the \href{https://chenjiahongbq.github.io/LCVSP}{\textcolor{magenta}{\textit{homepage}}}.
  \end{justify}
\end{figure*}

Table~\ref{tab:sota_3} reports the results on LINEMOD~\cite{hinterstoisser2012model}, which contains large viewpoint variations across object sequences.
Multi-reference methods can achieve high recalls, with FS6D+ICP~\cite{he2022fs6d} reaching 91.5\%, but they use multiple reference views and are not directly comparable to the single-reference setting.
Under the stricter single-reference setting, the single CAD-based FoundationPose~\cite{foundationposewen2024} obtains 48.3\%, while RGB-D image-based methods perform substantially better.
Our method achieves the best ADD(-S) among all single-reference methods, reaching 77.4\%.
It improves over CoordAR~\cite{zuo2026coordar}, SinRef-6D~\cite{liu2026scalable}, and UNOPose~\cite{Liu_2025_CVPR} by 2.4\%, 2.5\%, and 7.7\%, respectively.

\begin{table}[!ht]
	\centering
    \refstepcounter{table}
    \begin{justify}
    \small Table \Roman{table}: Pose estimation results on TUD-L~\cite{hodan2018bop} and YCB-V~\cite{posecnn} under the proposed challenging view-pair protocol. 
    We report the BOP average recall, AR$_{\text{BOP}}$ (\%).
    \end{justify}
    \label{tab:sota_our_tudl_ycbv}
    \begin{tabular*}{0.85\columnwidth}{@{\extracolsep{\fill}}lccc@{}}
        \toprule
        \multirow{2}{*}{Method} & \multicolumn{3}{c}{AR$_{\text{BOP}}$ (\%)} \\
        \cmidrule{2-4}
        & TUD-L~\cite{hodan2018bop} $\uparrow$ & YCB-V~\cite{posecnn} $\uparrow$ & Mean $\uparrow$ \\
        \midrule
        UNOPose~\cite{Liu_2025_CVPR} & \underline{55.9} & \underline{58.5} & \underline{57.2} \\
        CoordAR \cite{zuo2026coordar} & 36.7 & 50.2 & 43.5 \\
        SinRef-6D \cite{liu2026scalable} & 24.5 & 49.0 & 36.8 \\
		\midrule
		Ours & \textbf{67.0} & \textbf{64.2} & \textbf{65.6} \\
        \bottomrule
    \end{tabular*}
\end{table}

\textbf{Comparison under Our Predefined Challenging View Pair Protocol.} 
Table~\ref{tab:sota_our_tudl_ycbv} reports results under the proposed challenging view-pair protocol, which includes large viewpoint changes and perturbed object masks.
Our method achieves the best AR$_{\text{BOP}}$ on both datasets, with 67.0\% on TUD-L, 64.2\% on YCB-V, and 65.6\% on average.
Compared with the strongest baseline, UNOPose~\cite{Liu_2025_CVPR}, it improves AR by 11.1\% on TUD-L, 5.7\% on YCB-V, and 8.4\% on average.
This margin is larger than that under the standard UNOPose view-pair setting, where the viewpoint gap is more restricted.
These results highlight the advantage of our cross-view semantic prior when large viewpoint changes and mask noise make point-level matching ambiguous.

\subsection{Qualitative Results Analysis}
\label{sec:exp_qualitative}
\label{Qualitative}
We present extensive qualitative results, including pose estimation visual comparisons on each dataset, results under the challenging view pair protocol, and visualizations of cross-view interaction, to better analyze and demonstrate the superiority of our method.

\textbf{Qualitative Results on Six Datasets.}
Fig.~\ref{fig:comparison_vis} shows qualitative results on all six evaluated datasets, with object-level visualizations for each tested object.
These examples cover diverse challenges, including heavy occlusion, illumination variation, textureless objects, and geometric ambiguity.
Compared with UNOPose~\cite{Liu_2025_CVPR} and SinRef-6D~\cite{liu2026scalable}, our method produces more accurate pose alignment and lower object-level depth errors across different datasets.
\begin{figure}[ht]
	\centering
	\includegraphics[width=0.98\linewidth]{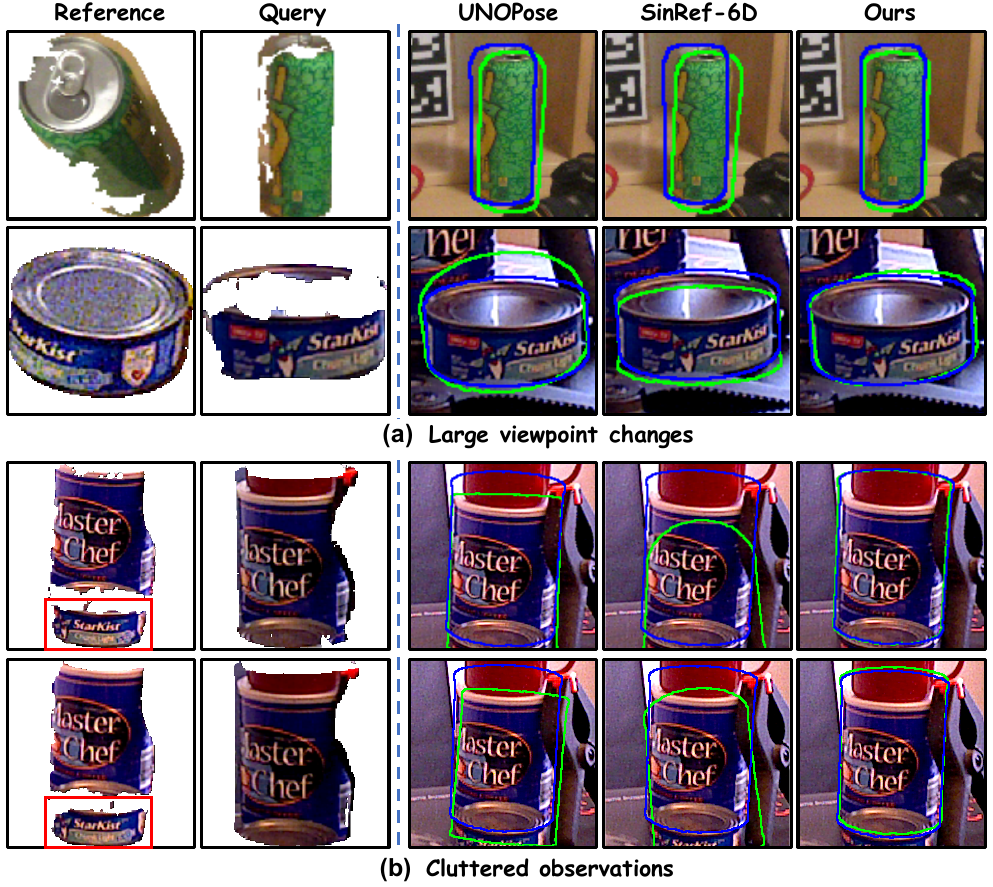}
    \refstepcounter{figure}
    \vspace{-20pt}
    \begin{justify}
        \small \textbf{Fig. 6: Qualitative comparison under the challenging view pair protocol.} 
        (a) Large viewpoint changes. 
        (b) Cluttered observations with heavy mask noise.
    \end{justify}
    \label{fig:vis_challenge}
\end{figure}

\textbf{Qualitative Results under the Challenging View Pair Protocol.}
Fig.~\ref{fig:vis_challenge} shows qualitative results under the challenging view-pair protocol, including large viewpoint changes and cluttered observations with heavy mask noise.
Under large viewpoint gaps, our projected contours remain better aligned with the ground truth than UNOPose~\cite{Liu_2025_CVPR} and SinRef-6D~\cite{liu2026scalable}, even when the visible overlap is limited.
Under heavy mask noise, nearby objects with similar geometry and appearance can introduce highly ambiguous correspondences.
Our method remains more stable in these cases because the cross-view semantic prior provides appearance, structure, and spatial guidance before geometry-aware matching.

\begin{figure}[!t]
  \centering
  \vspace{-15pt}
  \refstepcounter{figure}  \includegraphics[width=\linewidth]{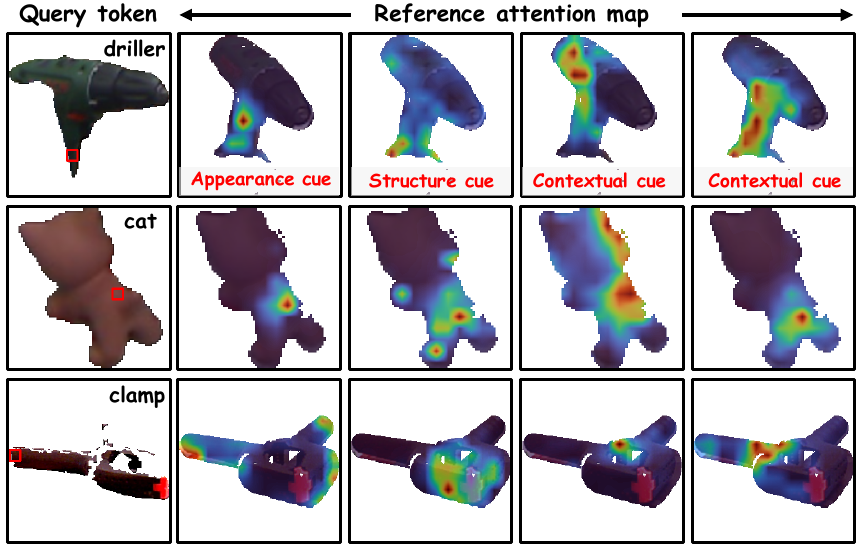}
  \label{fig:vis_attention_map}
  \vspace{-15pt}
  \begin{justify}
    \small \textbf{Fig. 7: Visualization of the attention maps in cross-view interaction.} We select one token in the query view (red box) and visualize the responses of different attention heads. 
  \end{justify}
\end{figure}

\textbf{Visualization of Cross-View Interaction.}
Fig.~\ref{fig:vis_attention_map} visualizes the attention responses from a selected query token to reference tokens in different CVSI heads.
Different heads attend to complementary reference regions, such as local parts, object boundaries, and broader body structures.
For example, in the driller case, the selected query token activates the handle area, the elongated body, and their junction.
Similar patterns appear in the cat and clamp examples, suggesting that CVSI aggregates appearance, structural, and contextual cues across views rather than simply averaging features.
These cross-view responses provide a semantic prior for point-level matching under viewpoint changes and local ambiguities.

\subsection{Ablation Studies}
\label{ablation}
\label{sec:exp_ablation}
We conduct ablation studies to verify the contribution of the proposed cross-view semantic prior and its training constraints.
We further examine key architectural choices and robustness under different reference viewpoint gaps.
Unless otherwise specified, all ablations are performed on YCB-V~\cite{posecnn}.

\textbf{Impact of Backbone Networks.}
We first study the influence of the VFM backbone to separate backbone improvements from the contribution of the proposed components.
As shown in Table~\ref{tab:ablation_main}, replacing DINOv2~\cite{oquab2023dinov2} with DINOv3~\cite{simeoni2025dinov3} improves the baseline AR$_{\text{BOP}}$ from 83.1\% to 84.2\% (A0 vs. B0).
This indicates that a stronger VFM backbone provides better visual representations for correspondence learning.
However, after introducing the proposed components, the gap between the two backbones becomes much smaller: the DINOv2 variant reaches 86.0\% (A1), while the corresponding DINOv3 variant reaches 86.2\% (B3).
This suggests that the performance gain is not mainly explained by the backbone replacement.
Since DINOv3 also uses a larger patch size than DINOv2 and is computationally lighter in our setting, we use DINOv3 as the default backbone in the remaining experiments.
\begin{table}[ht]
	\centering
    \refstepcounter{table}
    \begin{justify}
    \small Table \Roman{table}: Ablation study of the backbone and proposed components on the YCB-V~\cite{posecnn} dataset. 
	Results are reported with DINOv2~\cite{oquab2023dinov2} and DINOv3~\cite{simeoni2025dinov3} backbones. 
	(*) denotes the variant with shared cross-attention weights between the query and reference branches.
    \end{justify}
    \label{tab:ablation_main}
	\setlength{\tabcolsep}{3.7pt}
	\begin{tabular}{@{}ccccccccc@{}}
		\toprule
		Row & Backbone & CVSI & RAGC & IVSP & VSD & MSSD & MSPD & AR$_{\text{BOP}}$ \\
        \midrule
		A0 & \multirow{2}{*}{DINOv2} & \xmark & \xmark & \xmark & 82.6 & 87.4 & 79.4 & 83.1 \\
		A1 &  & \cmark & \cmark & \cmark & 83.0 & 90.4 & 84.5 & 86.0 \\
		\midrule
		B0 & \multirow{6}{*}{DINOv3} & \xmark & \xmark & \xmark & 82.4 & 88.4 & 81.7 & 84.2 \\
		B1 &  & \cmark & \xmark & \xmark & 82.9 & 90.1 & 84.0 & 85.7 \\
		B2 &  & \xmark & \cmark & \xmark & 83.0 & 89.0 & 82.2 & 84.7 \\
		B3 &  & \cmark & \cmark & \cmark & 83.5 & 90.7 & 84.5 & 86.2 \\
		B4 &  & \cmark & \cmark & \cmark & \textbf{83.9*} & \textbf{91.1*} & \textbf{85.1*} & \textbf{86.7*} \\
		B5 &  & \cmark & \cmark & \xmark & 82.9 & 90.2 & 84.1 & 85.7 \\
        \bottomrule
	\end{tabular}
\end{table}

\textbf{Effectiveness and Complementarity of CVSI, RAGC, and IVSP.} 
Table~\ref{tab:ablation_main} analyzes the contribution of the proposed components on YCB-V~\cite{posecnn}.
This ablation is designed to separate the effects of early cross-view semantic interaction, reference-anchored geometric supervision, and token structure preservation.

Under the DINOv3 backbone, adding only CVSI improves AR$_{\text{BOP}}$ from 84.2\% to 85.7\% (B1 vs. B0), giving a gain of 1.5\%.
The improvement is also consistent on MSSD and MSPD, which increase from 88.4\% to 90.1\% and from 81.7\% to 84.0\%, respectively.
This result shows that performing cross-view interaction before geometric decoding provides more discriminative features for correspondence estimation.
In particular, the larger gains on MSSD and MSPD suggest that CVSI improves both 3D surface alignment and projection accuracy, which are directly affected by correspondence quality.

The RAGC-only variant improves AR$_{\text{BOP}}$ from 84.2\% to 84.7\% (B2 vs. B0).
This moderate gain indicates that reference-anchored geometric supervision provides a useful geometric regularization signal, but it is not the main source of the overall improvement.
This comparison is important because it controls for the effect of the auxiliary coordinate prediction head: adding the RAGC head and loss alone cannot explain the full gain of the proposed method.
Instead, RAGC is most useful when it works with cross-view semantic conditioning, where it encourages the decoded point features to be consistent in a shared 3D reference frame.

The role of IVSP is also clarified by comparing B1, B3, and B5.
When CVSI and RAGC are used without IVSP, the AR$_{\text{BOP}}$ remains 85.7\% (B5), the same as CVSI alone (B1).
After adding IVSP, the performance increases to 86.2\% (B3), with consistent gains on VSD, MSSD, and MSPD.
This shows that IVSP is not merely an additional auxiliary loss, but a stabilizing regularizer for cross-view token interaction.
By preserving the intra-view token affinity structure, IVSP prevents the interacted tokens from losing the original VFM organization, allowing the semantic prior introduced by CVSI and the geometric constraint imposed by RAGC to better complement each other.

Finally, sharing the cross-attention weights between the query and reference branches further improves AR$_{\text{BOP}}$ from 86.2\% to 86.7\% (B4 vs. B3).
This shared design encourages the two branches to follow a consistent interaction pattern and reduces redundant parameters.
Compared with the DINOv3 baseline, the full model improves AR$_{\text{BOP}}$ by 2.5\%, from 84.2\% to 86.7\%.
The improvements are also observed across all three BOP metrics, with gains of 1.5, 2.7, and 3.4\% on VSD, MSSD, and MSPD, respectively.
These results support the complementary design of the proposed framework: CVSI provides the main cross-view semantic prior, RAGC grounds the decoded features in 3D reference coordinates, and IVSP stabilizes the token structure during interaction.
Since RAGC and IVSP are used only during training, they improve feature learning without adding inference time computation.
We therefore adopt the shared CVSI design with both RAGC and IVSP as the default configuration.

\begin{figure}[!t]
	\centering
    \refstepcounter{figure}
	\includegraphics[width=\linewidth]{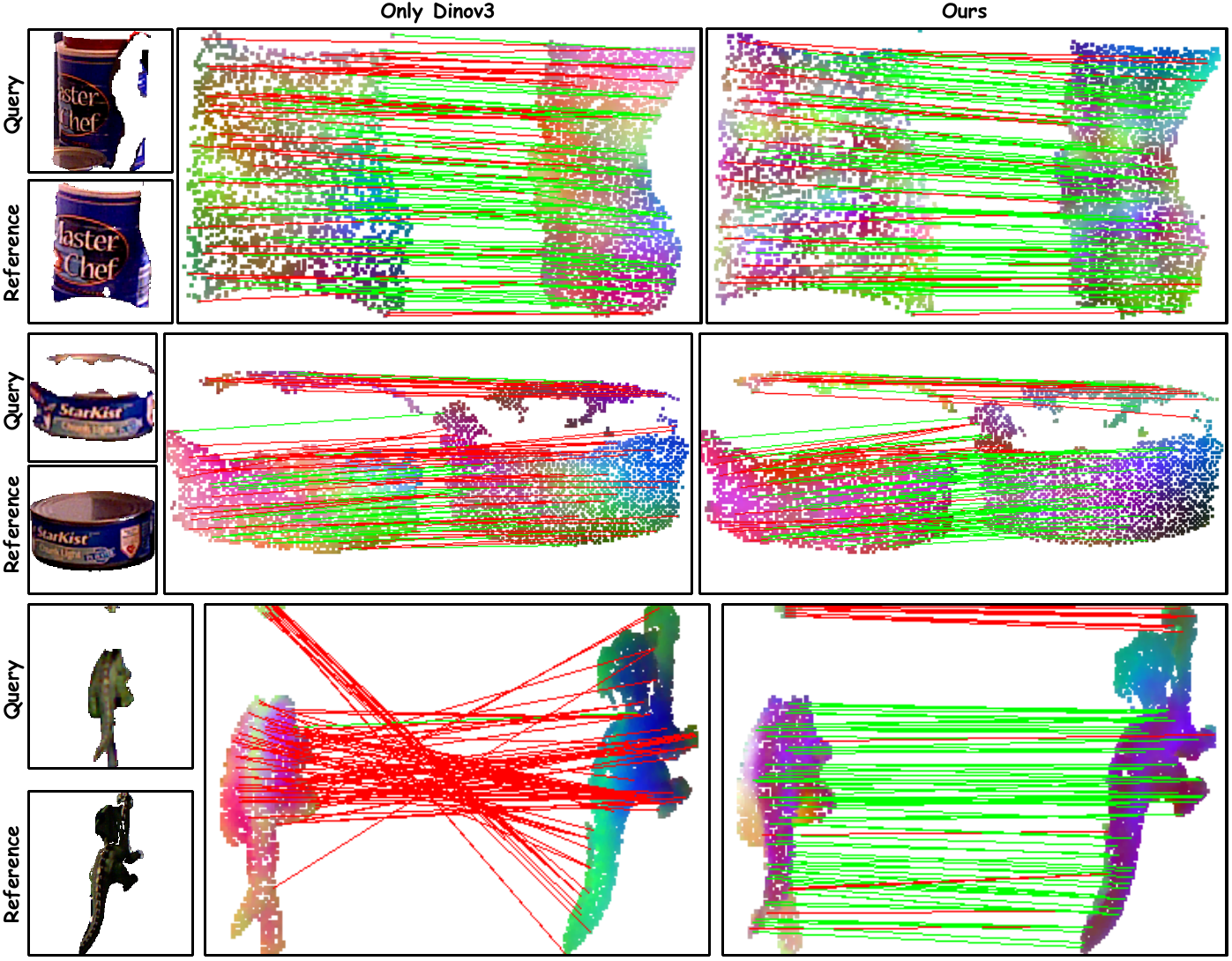}
    \label{fig:ablation_correspondences}
    \vspace{-15pt}
    \begin{justify}
        \small \textbf{Fig. 8: Visualization of correspondence estimation.} We compare the predicted correspondences obtained using only the DINOv3~\cite{simeoni2025dinov3} backbone and our full method. 
        Green and red lines denote geometrically consistent and inconsistent correspondences, respectively.
    \end{justify}
\end{figure}

\textbf{Visualization of Correspondences.}
Fig.~\ref{fig:ablation_correspondences} visualizes the correspondences produced by the DINOv3-only baseline and our full model.
The DINOv3-only baseline produces scattered matches, often connecting non-corresponding or locally similar object parts under partial overlap and noisy observations.
In contrast, our method yields more spatially coherent correspondences concentrated on geometrically compatible regions.
This indicates that the cross-view semantic prior improves point-feature discriminability before matching and provides cleaner correspondences for subsequent weighted SVD pose estimation.

\begin{figure}[!t]
	\centering
    \refstepcounter{figure}
	\includegraphics[width=\linewidth]{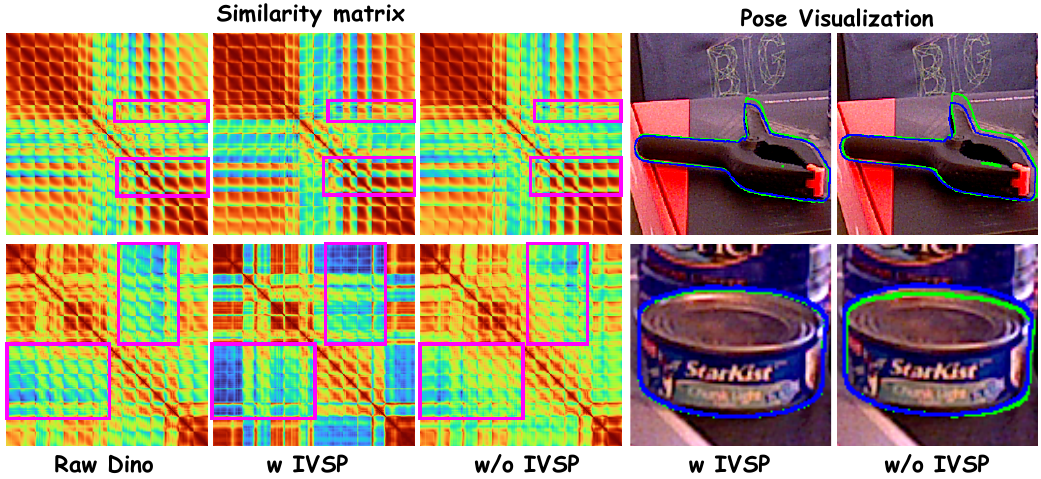}
    \label{fig:isp_vis}
    \vspace{-15pt}
    \begin{justify}
        \small \textbf{Fig. 9: Effectivenes of IVSP loss.} Without IVSP, cross-view interaction over-smooths the intra-view similarity structure, weakening local part and boundary contrast. 
        With IVSP, the interacted features preserve sharper relative similarity patterns inherited from DINO features.
        Best viewed when zoomed in.
    \end{justify}
\end{figure}

\textbf{Intra-View Similarity Structure Analysis.}
Fig.~\ref{fig:isp_vis} compares the intra-view token similarity matrices before and after cross-view interaction.
Raw DINO features show clear local affinity structures related to object parts, boundaries, and spatial layouts.
Without IVSP, cross-view interaction tends to over-smooth these structures, weakening the contrast of the highlighted local regions.
With IVSP, the interacted tokens preserve the major similarity topology of the original DINO features while keeping local regions more distinguishable.
This suggests that IVSP stabilizes CVSI by preventing excessive feature mixing, rather than simply suppressing cross-view adaptation.
The corresponding pose visualizations further show that preserving such token structure leads to more accurate contour alignment.

\textbf{Number of Cross-View Semantic Interaction Layers.}
Table~\ref{tab:number_CVSI} studies the effect of the number of CVSI layers.
Removing token-level cross-view interaction ($L=0$) gives 84.7\% AR$_{\text{BOP}}$, while adding one CVSI layer improves it to 85.7\%.
Increasing the depth further brings the best result at $L=3$, reaching 86.2\% AR$_{\text{BOP}}$.
Compared with $L=0$, $L=3$ improves VSD, MSSD, MSPD, and AR$_{\text{BOP}}$ by 0.5\%, 1.7\%, 2.3\%, and 1.5\%, respectively, showing that repeated cross-view interaction refines semantic context for correspondence learning.
We therefore use $L=3$ as the default setting.
\begin{table}[ht]
	\centering
    \refstepcounter{table}
    \begin{justify}
    \small Table \Roman{table}: Pose estimation performance under different numbers of CVSI layers on YCB-V~\cite{posecnn}. 
    The study is based on the B3 setting in Table~\ref{tab:ablation_main}, with $L=0$ denoting the variant without token-level cross-view interaction.
    \end{justify}
    \label{tab:number_CVSI}
    \setlength{\tabcolsep}{6.5pt}
    \begin{tabular}{@{}ccccc@{}}
    \toprule
    Number of $L$ & VSD & MSSD & MSPD & AR$_{\text{BOP}}$ \\
    \midrule
    0 & 83.0 & 89.0 & 82.2 & 84.7 \\
    1 & 82.8 & 90.2 & 84.0 & 85.7 \\
    2 & 83.1 & 90.2 & 84.0 & 85.8 \\
    3 & \textbf{83.5} & \textbf{90.7} & \textbf{84.5} & \textbf{86.2} \\
    \bottomrule
	\end{tabular}
\end{table}

\begin{table*}[!t]
	\centering
    \refstepcounter{table}
    \begin{justify}
    \small Table \Roman{table}: Pose estimation performance under different geometric decoder~\cite{qin2023geotransformer} depths on YCB-V~\cite{posecnn}. 
	The column ``CVSI Prior'' denotes whether the dense cross-view semantic prior is introduced before geometric decoding. 
    H1 is a geometry only control with a parameter scale comparable to our default model. 
    The runtime includes only pose estimation and excludes segmentation. 
    \end{justify}
    \label{tab:number_Geotrans}
	\begin{tabular*}{0.83\textwidth}{@{\extracolsep{\fill}}cccccccccc@{}}
		\toprule
		Row & Geo. Dec. Layers & CVSI Prior & VSD & MSSD & MSPD & AR$_{\text{BOP}}$ & Train Params (M) & GFLOPs & Runtime (s) \\
        \midrule
		C0 & 1 & \cmark & 81.9 & 89.7 & 82.3 & 84.6 & 23.35 & 193.79 & 0.625 \\
		D0 & 2 & \cmark & 82.5 & 89.7 & 83.0 & 85.1 & 26.12 & 221.41 & 0.662 \\
		\midrule
		E0 & \multirow{2}{*}{3} & \xmark & 82.4 & 88.4 & 81.7 & 84.2 & 21.37 & 233.61 & 0.686 \\
		E1 &  & \cmark & \textbf{83.9} & \textbf{91.1} & \textbf{85.1} & \textbf{86.7} & 28.88 & 249.02 & 0.711 \\
		\midrule
		F0 & \multirow{2}{*}{4} & \xmark & 82.5 & 88.5 & 82.0 & 84.3 & 24.14 & 261.23 & 0.721 \\
		F1 &  & \cmark & 83.3 & 90.7 & 84.8 & 86.3 & 31.65 & 276.64 & 0.789 \\
		\midrule
		G0 & \multirow{2}{*}{5} & \xmark & 82.6 & 88.8 & 82.4 & 84.6 & 26.91 & 288.85 & 0.777 \\
		G1 &  & \cmark & 83.8 & 90.8 & 84.9 & 86.5 & 34.42 & 304.26 & 0.794 \\
		\midrule
		H1 & 6 & \xmark & 82.6 & 88.5 & 82.4 & 84.5 & 29.68 & 316.46 & 0.803 \\
        \bottomrule
	\end{tabular*}
\end{table*}

\textbf{Geometric Decoder Depth and Cross-View Semantic Prior.}
Table~\ref{tab:number_Geotrans} studies the relation between geometric decoder depth and the proposed CVSI semantic prior.
This ablation aims to examine whether stronger geometric decoding alone can replace the early cross-view semantic prior.

We first evaluate the effect of increasing the geometric decoder depth when the CVSI prior is used.
Increasing the decoder from 1 to 3 layers improves AR$_{\text{BOP}}$ from 84.6\% to 86.7\% (C0, D0, and E1), showing that geometric decoding is important for fusing semantic and geometric cues into correspondence features.
However, further increasing the decoder from 3 to 5 layers slightly decreases AR$_{\text{BOP}}$ from 86.7\% to 86.5\% (E1 vs. G1), while increasing trainable parameters, FLOPs, and runtime.
This suggests that 3 decoder layers are sufficient once the point features are guided by the CVSI prior.

We then remove the CVSI prior and increase only the geometric decoder depth.
Without CVSI, increasing the decoder from 3 to 5 layers improves AR$_{\text{BOP}}$ only marginally from 84.2\% to 84.6\% (E0 vs. G0), while runtime increases from 0.686\,s to 0.777\,s.
The geometry only control H1 further increases the trainable parameters to 29.68M and the computation to 316.46 GFLOPs, but still obtains only 84.5\% AR$_{\text{BOP}}$.
In contrast, our 3-layer decoder with the CVSI prior achieves 86.7\% AR$_{\text{BOP}}$ with 28.88M trainable parameters and 249.02 GFLOPs.
Compared with H1, our model improves AR$_{\text{BOP}}$ by 2.2\% while using fewer parameters and 21.3\% fewer FLOPs.
Compared with the 5-layer geometry only baseline G0, our model improves AR$_{\text{BOP}}$ by 2.1\% with 13.8\% fewer FLOPs and lower runtime.

These results indicate that the improvement is not mainly caused by decoder depth, parameter count, or computation.
Instead, geometric decoding provides 3D consistency, while the CVSI prior supplies cross-view semantic guidance before decoding, enabling the point features to better combine semantic discriminability with geometric consistency.
\begin{figure}[!ht]
	\centering
    \refstepcounter{figure}
	\includegraphics[width=\linewidth]{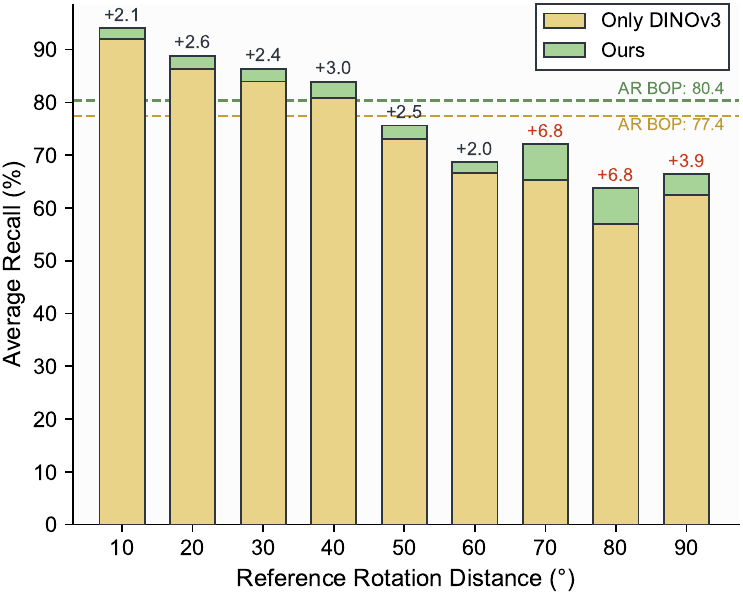}
    \label{fig:ablation_rotation_distance}
    \vspace{-15pt}
    \begin{justify}
        \small \textbf{Fig. 10: Effect of the reference viewpoint gap on YCB-V~\cite{posecnn}.} We compare the performance of using only the DINOv3~\cite{simeoni2025dinov3} backbone and our full method. We divide the rotation angles into nine groups. 
        Our method consistently improves AR$_{\text{BOP}}$ across all bins, with larger gains under severe viewpoint changes.
    \end{justify}
\end{figure}

\textbf{Effect of Reference Viewpoint Gap.}
To evaluate robustness under different viewpoint gaps, we randomly select reference views on YCB-V~\cite{posecnn} with relative rotations from 0 to 90 degrees and divide the results into nine bins.
As shown in Fig.~\ref{fig:ablation_rotation_distance}, our method consistently outperforms the DINOv3-only baseline across all rotation ranges, improving the overall AR$_{\text{BOP}}$ from 77.4\% to 80.4\%.
The gain is moderate from 10 to 60 degrees, but becomes larger under severe viewpoint changes, reaching 6.8\% at both 70 and 80 degrees and 3.9\% at 90 degrees.
This trend shows that the cross-view semantic prior is particularly beneficial when valid overlap becomes sparse and ambiguous.

\subsection{Failure cases}
\label{sec:exp_failure}

\begin{figure}[!t]
  \centering
  \refstepcounter{figure}  \includegraphics[width=\linewidth]{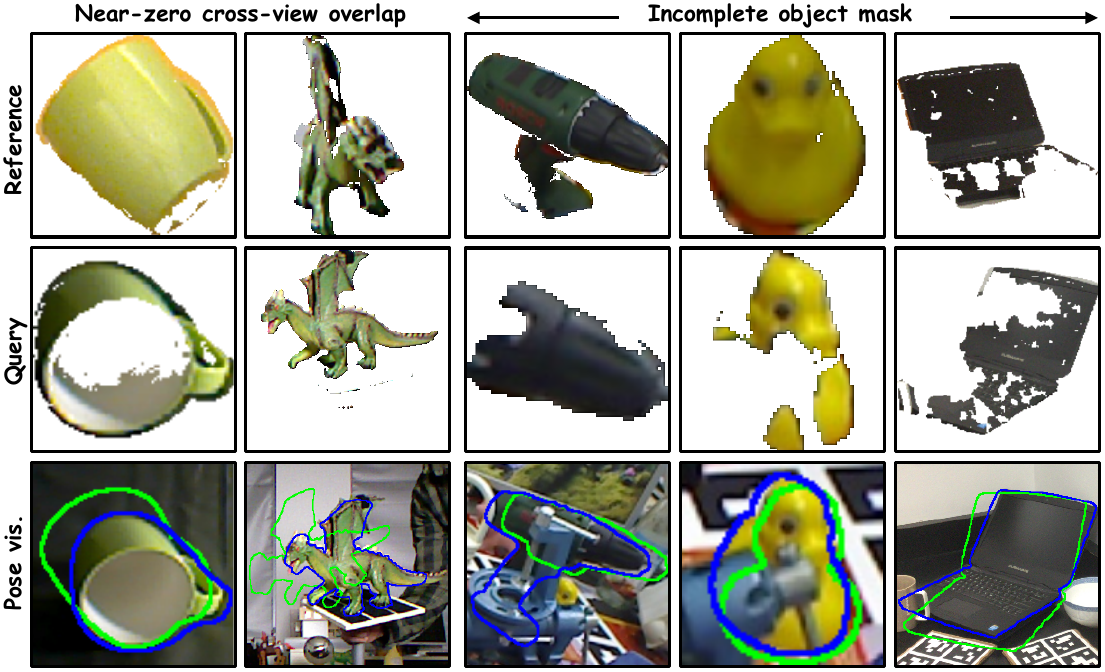}
  \label{fig:failure_cases}
  \vspace{-15pt}
  \begin{justify}
    \small \textbf{Fig. 11: Failure cases under challenging scenes across different datasets.} These examples include cases with nearly zero cross-view overlap and incomplete object masks.
  \end{justify}
\end{figure}

Although our method improves correspondence reliability under large viewpoint changes and mask noise, it is still subject to the observability limits of single-reference correspondence-based pose estimation.
As shown in Fig.~\ref{fig:failure_cases}, failures mainly occur in two cases: near-zero visible overlap between the query and reference views, and highly incomplete object masks.
In the first case, too few shared regions are available to support valid cross-view correspondences.
In the second case, missing discriminative object parts provide insufficient evidence for both token-level semantic interaction and point-level geometric matching.
These cases represent boundary conditions of the single-reference correspondence formulation.

\section{CONCLUSIONS}
\label{sec:conclusion}
This paper studied single-reference unseen object 6D pose estimation and identified the insufficient cross-view exchange of dense visual-semantic cues as a key limitation of existing correspondence-based pipelines.
To address this issue, we proposed a cross-view semantic prior learning framework that introduces dense token-level CVSI before geometric decoding, preserves intra-view token structure with IVSP, and grounds decoded features through RAGC.
Together, these components improve the semantic and geometric discriminability of correspondence features for more reliable point cloud matching and weighted SVD pose estimation.
Extensive experiments on six benchmark datasets and a challenging view-pair protocol demonstrate state-of-the-art performance with comparable inference speed.

\textbf{Limitations and Future Work.}
Despite these improvements, the proposed method remains limited by the input evidence available in the single-reference setting.
When the query and reference views have near-zero overlap or the masks remove discriminative object regions, reliable correspondences remain difficult to establish.
Future work will extend the framework to multiple query observations and explicit multi-view geometric reasoning, which may provide richer object evidence and alleviate these failure cases.

\renewcommand*{\bibfont}{\footnotesize}
\printbibliography

\end{document}